\documentclass[10pt,twocolumn,letterpaper]{article}

\usepackage{iccv}
\usepackage{times}
\usepackage{epsfig}
\usepackage{graphicx}
\usepackage{amsmath}
\usepackage{amsthm}
\usepackage{amssymb}
\usepackage{subfigure}
\usepackage{algorithm}
\usepackage{algorithmic}
\usepackage{threeparttable}
\usepackage{booktabs}
\usepackage{multirow}
\usepackage[misc]{ifsym} 

\newtheorem{lemma}{Lemma}
\newtheorem{claim}{Claim}

\DeclareMathOperator*{\argmax}{arg\,max}

\usepackage[breaklinks=true,bookmarks=false]{hyperref}

\iccvfinalcopy % *** Uncomment this line for the final submission

\def\iccvPaperID{****} % *** Enter the ICCV Paper ID here

\newcommand*{\affmark}[1][*]{\textsuperscript{#1}}

\hypersetup{
	colorlinks=true,
	linkcolor=red,
}
\begin{document}

%%%%%%%%% TITLE
\title{Deep Comprehensive Correlation Mining for Image Clustering}
\author{
	Jianlong Wu$^{123}\thanks{\scriptsize{Equal contribution and the work was done during interns at SenseTime Research}}$ \ \  Keyu Long$^{2*}$ \  Fei Wang$^{2}$ \ Chen Qian$^{2}$ \  Cheng Li$^{2}$ \  Zhouchen Lin$^{3}$\affmark[(\Letter)] \ Hongbin Zha$^{3}$ \\
	$^{1}$School of Computer Science and Technology, Shandong University \\
	$^{2}$SenseTime Research \\
	$^{3}$Key Laboratory of Machine Perception (MOE), School of EECS, Peking University\\
%	{\small jlwu1992@pku.edu.cn, keyull@163.com,cory$\_$longkeyu@outlook.com, \{wangfei, qianchen, chengli\}@sensetime.com, zlin@pku.edu.cn, zha@cis.pku.edu.cn }
	{\small jlwu1992@sdu.edu.cn, corylky114@gmail.com, \{wangfei, qianchen, chengli\}@sensetime.com, zlin@pku.edu.cn, zha@cis.pku.edu.cn }
}

%\title{Deep Comprehensive Correlation Mining for Image Clustering}
%\author{
%	Jianlong Wu \quad Keyu Long \quad Fei Wang \quad Chen Qian \quad Cheng Li \quad Zhouchen Lin \quad Hongbin Zha \vspace{3mm}\\
%	Key Laboratory of Machine Perception (MOE), School of EECS, Peking University, Beijing 100871, P. R. China \\
%	Sensetime \\
%	Cooperative Medianet Innovation Center, Shanghai Jiao Tong University, Shanghai 200240, P. R. China
%}

%\and

\maketitle
%\thispagestyle{empty}

%%%%%%%%% ABSTRACT
\begin{abstract}
	
	Recent developed deep unsupervised methods allow us to jointly learn representation and cluster unlabelled data. 
	These deep clustering methods %like DAC start with 
	mainly focus on the correlation among samples, e.g., 
	selecting high precision pairs to gradually tune the feature representation, which neglects other useful correlations.
	In this paper, we propose a novel clustering framework, named deep comprehensive correlation mining~(DCCM), for exploring and taking full advantage of various kinds of correlations behind the unlabeled data from three aspects: 
	1) Instead of only using pair-wise information, pseudo-label supervision is proposed to investigate category information and learn discriminative features.
	2) The features' robustness to image transformation of input space is fully explored, which benefits the network learning and significantly improves the performance.
	3) The triplet mutual information among features is presented for clustering problem to lift the recently discovered instance-level deep mutual information to a triplet-level formation, which further helps to learn more discriminative features. Extensive experiments on several challenging datasets show that our method achieves good performance, e.g., attaining $62.3\%$ clustering accuracy on CIFAR-10, which is $10.1\%$ higher than the state-of-the-art results\footnote{\scriptsize{Project address: \url{https://github.com/Cory-M/DCCM}}}.
	%https://github.com/Cory-M/DCCM
	%Extensive experiments on several challenging datasets show that our method achieves good performance, e.g., attaining $62.3\%$ clustering accuracy on CIFAR-10, and $34.0\%$ on CIFAR-100, both of which significantly surpass the state-of-the-art results more than $10.0\%$\footnote{Project address: \url{https://jlwu1992.github.io/DCCM/}}.
	% Instead of only using pair-wise information, we propose the pseudo-label loss
	%, a more powerful loss to learn discriminative feature between categories, which leads to a better convergence and better clustering results. 
	%2) We find that considering features' invariance to image transformation will significantly improve the performance. 
	%3) We extend recently discovered deep mutual information into a triplet formation in clustering problem, which further helps to learn more discriminative features. Extensive experiments on several challenging datasets demonstrate that our method achieves superior performance over other state-of-the-art methods, e.g., attaining $62.3\%$ clustering accuracy on CIFAR-10, and $34.0\%$ on CIFAR-100.
	
\end{abstract}

%%%%%%%%% BODY TEXT
%------------------------------------------------------------------ 
\vspace{-3mm}
\section{Introduction}
%------------------------------------------------------------------ 
Clustering is one of the fundamental tasks in computer vision and machine learning. 
Especially with the development of the Internet, we can easily collect thousands of images and videos every day, most of which are unlabeled.
It is very expensive and time-consuming to manually label these data. 
%However, most existing approaches, especially supervised deep learning, heavily rely on label information.
In order to make use of these unlabeled data and investigate their correlations,
unsupervised clustering draws much attention recently, which aims to categorize similar data into one cluster based on some similarity measures.
\par
%------------------------------------------------------------------
\begin{figure}[t]
	\centering 
	\includegraphics[width=0.83\linewidth]{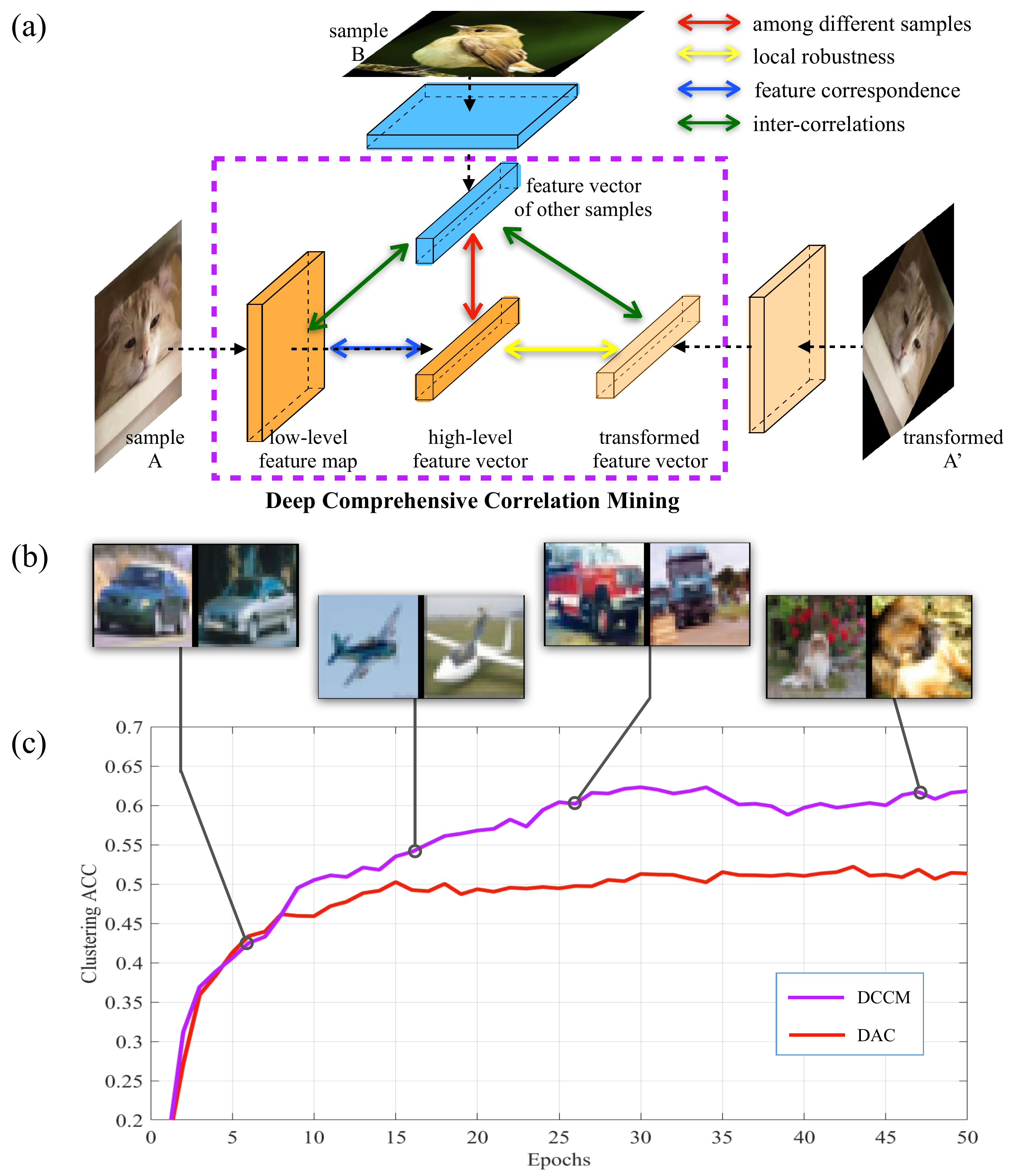}
	%	\caption{Comprehensive correlations mined in this paper and superiority over DAC. (a) Various correlations between samples
	%	Lines in different color denote various correlations; (b) we can learn more discriminative feature in a progressive manner; (c) DCCM achieves much better results than DAC.
	%	} 
	\caption{Comprehensive correlations mining. (a) Various correlations; (b) Connect pair-wise items in higher semantic level progressively; (c) Better results of DCCM than the state-of-the-art DAC~\cite{chang2017deep} on CIFAR-10~\cite{cifar10}. Best viewed in color!
	} 
    \vspace{-5mm}
	\label{fig:motivation}
\end{figure}
%------------------------------------------------------------------
%------------------------------------------------------------------
%\begin{figure}[t]
%	\centering 
%	\includegraphics[width=0.7\linewidth]{Figure_1.png}
%	\caption{Clustering ACC comparison with DAC on CIFAR-10.} 
%	\vspace{-5mm}%
%	\label{fig:ACC_DAC}
%\end{figure}
%------------------------------------------------------------------
%For natural images, the shape, appearance, and color may vary from sample to sample, even they share the same semantic meaning.
%
Image clustering is a challenging task due to the image variance of shape and appearance in the wild.
%So image clustering is a very challenging task.
Traditional clustering methods~\cite{zelnik2005self, gowda1978agglomerative, cai2009locality}, such as K-means, spectral clustering~\cite{ng2002spectral,8736495}, and subspace clustering~\cite{liu2013robust, elhamifar2009sparse} may fail for two main issues:
first, hand-crafted features have limited capacity and cannot dynamically adjust to capture the prior distribution, especially when dealing with large-scale real-world images;
second, the separation of feature extraction and clustering will make the solution sub-optimal.
Recently, with the booming of deep learning~\cite{krizhevsky2012imagenet,he2016deep,wang2017residual,li2018recurrent,xie2019differentiable}, many researchers shift their attention to deep unsupervised feature learning and clustering~\cite{peng2016deep, ji2017deep,chang2017deep}, which can well solve the aforementioned limitations.
Typically, to learn a better representation, \cite{bengio2007greedy,2016ICMLDEC,yang2017towards} adopt the auto-encoder and~\cite{hjelm2018learning} maximizes the mutual information between features. 
DAC~\cite{chang2017deep} constructs positive and negative pairs to guide network training.

However, for these methods, several points are still missing.
Firstly, feature representations that only consider reconstruction or mutual information lack discriminative power.
Secondly, traditional cluster method like K-means effectively use category assumption on data. Contrast to that, DAC only focuses on pair-wise correlation and neglects the category information, which limits its performance.
Thirdly, there are also other correlations that are helpful for deep image feature learning, for example, \cite{lenc2015understanding} shows that measuring feature equivariance can benefit image representation understanding.
\par
To tackle above issues, as shown in Figure~\ref{fig:motivation}(a), we propose a novel method, namely deep comprehensive correlation mining~(DCCM), which comprehensively explores correlations among different samples~(red line), local robustness to geometry transformation~(yellow line), between different layer features of the same sample~(blue line), and their inter-correlations~(green lines) to learn discriminative representations and train the network in a progressive manner.
First of all, for the correlation among different samples, we adopt the deep convolutional neural network~(CNN) to generate prediction feature for the input image. With proper constraints, the learned prediction feature will tend to be one-hot.
Then we can compute the cosine similarity and construct the similarity graph.
%construct the similarity graph by computing the similarity   
%we follow the basic pipeline of dac to use graph information to generate pseudo-graph as supervision. One-hot label. We also generate high-confident pseudo-label to guide the network training.
Based on the similarity graph and prediction feature, 
we assign a large threshold to get highly-confident pseudo-graph and pseudo-label to guide the feature learning.
Secondly, for the local robustness to small perturbations, we add small perturbation or transformation on the original input image to generate a transformed image.
Under the local robustness assumption, the prediction of the transformed image should be consistent with that of the original image. So we can use the prediction of the original image to guide the feature learning of the transformed image.
Thirdly, feature representation of deep layer should preserve distinct information of the input.
So we maximize the mutual information between the deep layer feature and shallow layer feature of the same sample. 
To make the representation more discriminative, we further extend it to a triplet form by incorporating the graph information above.
Finally, we combine the loss function of these three different aspects and 
jointly investigate these correlations in an end-to-end way.
Results in Figure~\ref{fig:motivation}(c) show the superiority of our method~(purple curve) over the state-of-the-art method DAC~\cite{chang2017deep}~(red curve).

Our main contributions are summarized as follows: 
\vspace{-1mm}
\begin{itemize}
	\setlength{\itemsep}{-0.5pt}
	\item[1)] We propose a novel end-to-end deep clustering framework to comprehensively mine various kinds of correlations, and select highly-confident information to train the network in a progressive way; 
	\item[2)] We first derive the rationality of pseudo-label and introduce the highly-confident pseudo-label loss to directly investigate the category information and guide the unsupervised training of deep network;
	\item[3)] We make use of the local robustness assumption and utilize above pseudo-graph and pseudo-label to learn better representation;
	%Instead of simply constrain the feature distance, above pseudo-graph and pseudo-label is utilized to guide discriminative feature learning of samples after small perturbation;
	\item[4)] We extend the instance-level mutual information to triplet-level, and come up with triplet mutual information loss to learn more discriminative features.
	%mutual information by constructing more positive and negatives pairs based on the pseudo-graph information to learn more discriminative features.
	%Mutual information between deep layer feature and shallow layer feature of a same sample is maximized. 
	%\vspace{-3mm}
	%\item[d)] The proposed method achieves very good performance on several challenging real-world datasets.
\end{itemize}
%------------------------------------------------------------------ 
\vspace{-2mm}
%------------------------------------------------------------------ 
\section{Related Work}
\vspace{-1mm}
%------------------------------------------------------------------ 
\subsection{Deep Clustering}
%------------------------------------------------------------------ 
Existing deep clustering methods~\cite{yang2016joint,2016ICMLDEC,chang2017deep} mainly aim to combine the deep feature learning~\cite{bengio2007greedy,vincent2010stacked,zeiler2010deconvolutional} with traditional clustering methods~\cite{zelnik2005self,gowda1978agglomerative,cai2009locality}.
%One typical way is to minimize the loss of traditional clustering methods to guide the feature learning of deep neural network.
Auto-encoder~(AE)~\cite{bengio2007greedy} is a very popular feature learning method for deep clustering, and many methods are proposed to minimize the loss of traditional clustering methods to regularize the learning of latent representation of auto-encoder.
For example,
\cite{2016ICMLDEC, guo2017improved} proposes the deep embedding clustering to utilize the KL-divergence loss.
\cite{depict} also uses the KL-divergence loss, but adds a noisy encoder to learn more robust representation.
\cite{yang2017towards} adopts the K-means loss, and
\cite{ji2017deep,peng2016deep, xi2017cascade} incorporate the self-representation based subspace clustering loss.
\par
Besides the auto-encoder, some methods directly design specific loss function based on the last layer output.
\cite{yang2016joint} introduces a recurrent-agglomerative framework to merge clusters that are close to each other.
\cite{chang2017deep} explores the correlation among different samples based on the label features, and uses such similarity as supervision.
\cite{shaham2018spectralnet} extends the spectral clustering into deep formulation.
%------------------------------------------------------------------ 
\subsection{Deep Unsupervised Feature Learning}
%------------------------------------------------------------------ 
Instead of clustering, several approaches~\cite{bengio2007greedy,kingma2013auto, makhzani2015adversarial,dosovitskiy2015discriminative,oyallon2015deep,bautista2016cliquecnn,wang2017unsupervised,wu2018unsupervised} mainly focus on deep unsupervised learning of representations.
Based on Generative Adversarial Networks~(GAN), \cite{donahue2016adversarial} proposes to add an encoder to extract visual features.
\cite{bojanowski2017unsupervised} directly uses the fixed targets which are uniformly sampled from a unit sphere to constrain the deep features assignment.
\cite{caron2018deep} utilizes the pseudo-label computed by the K-means on output features as supervision to train the deep neural networks.
\cite{hjelm2018learning} proposes the deep infomax to maximize the mutual information between the input and output of a deep neural network encoder.
%------------------------------------------------------------------ 
\subsection{Self-supervised Learning}
%------------------------------------------------------------------ 
Self-supervised learning~\cite{jing2019self, kolesnikov2019revisiting} generally needs to design a pretext task, where a target objective can be computed without supervision.
They assume that the learned representations of the pretext task contain high-level semantic information that is useful for solving downstream tasks of interest, such as image classification.
For example, 
\cite{doersch2015unsupervised} tries to predict the relative location of image patches, and 
\cite{noroozi2016unsupervised, noroozi2018boosting} predict the permutation of a “jigsaw puzzle” created from the full image.
\cite{dosovitskiy2014discriminative} regards each image as an individual class and generates multiple images of it by data augmentation to train the network.
\cite{gidaris2018unsupervised} rotates
an image randomly by one of four different angles and lets the deep model
predict the rotation.
\vspace{-1mm}
%------------------------------------------------------------------ 
\section{Deep Comprehensive Correlation Mining}
\vspace{-1mm}
%------------------------------------------------------------------
Without labels, correlation stands in the most important place in deep clustering.
In this section, we first construct pseudo-graph to explore binary correlation between samples to start the network training.
Then we propose the pseudo-label loss to make full use of category information behind the data.
Next, we mine the local robustness of predictions before and after adding transform on input image.
We also lift the instance level mutual information to triplet level to make it more discriminative.
Finally, we combine them together to get our proposed method.
%We mainly consider three different kinds of correlations for deep image clustering, including: 
%1) the correlation among different samples,
%2) robustness of predictions before and after adding transform on input image, and  3) feature correspondence between deep and shallow layer features of the same sample.
%Their inter-correlations are also well investigated.
%In this section, 
%we first design model to achieve the goal of each kind of correlation investigation,
%and then unify them together to get our proposed method.
%------------------------------------------------------------------ 
\subsection{Preliminary: Pseudo-graph Supervision}\label{subsection:pg}
%------------------------------------------------------------------ 
%------------------------------------------------------------------
\begin{figure*}[!ht]
	\centering 
	\includegraphics[width=0.93\linewidth]{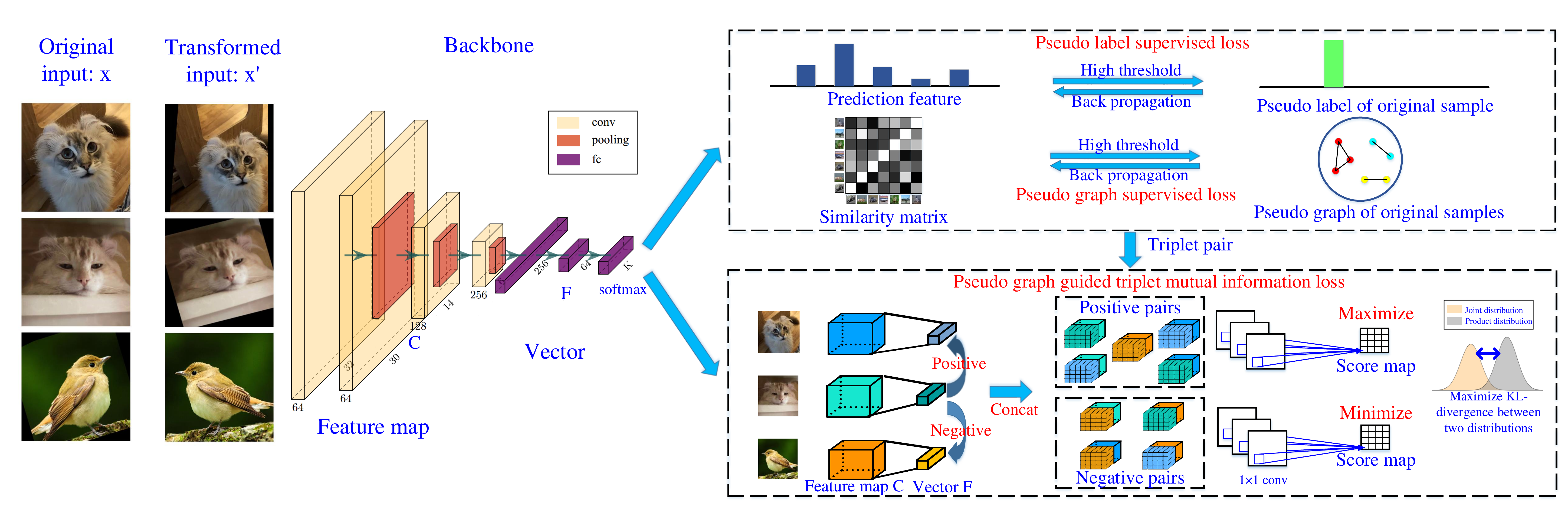}
	\caption{The pipeline of the proposed DCCM method. Based on the ideally one-hot prediction feature, we compute the highly-confident pseudo-graph and pseudo-label to guide the feature learning of both original and transformed samples, investigating both correlations among different samples and local robustness after small perturbation. Meanwhile, to investigate discriminative feature correspondence, the pseudo-graph is utilized to select highly-confident positive and negative pairs for triplet mutual information optimization.
	} 
	\label{fig:pipline}
	\vspace{-4mm}
\end{figure*}
%------------------------------------------------------------------
We first compute the similarity among samples and select highly-confident pair-wise information to guide the network training by constructing pseudo-graph.
Let $\boldsymbol{\mathcal{X}} = \{ \mathbf{x}_{i} \}_{i=1}^{N}$ be the unlabeled dataset, where $\mathbf{x}_{i}$ is the $i$-th image and $N$ is the total number of images. Denote  $K$ as the total number of classes.
We aim to learn a deep CNN based mapping function $f$ which is parameterized by $\theta$.
Then we can use $\mathbf{z}_{i} = f_{\theta}(\mathbf{x}_{i})\in \mathbb{R}^{K}$ to represent the prediction feature of image $\mathbf{x}_i$ after the softmax layer of CNN.
It has the following properties:
\vspace{-1mm}
\begin{equation}
	\sum_{t=1}^{K} z_{it} \!   =  \! 1, \forall i \! = \! 1,\cdots, N, \text{and} \ z_{it}  \! \geq  \! 0, \forall t \! = \! 1,\cdots,K. 
	\label{eq:z_property}
	\vspace{-1mm}
\end{equation}
Based on the label feature $\mathbf{z}$, the cosine similarity  between the $i$-th and the $j$-th samples can be computed by 
$S_{ij} = \frac{\mathbf{z}_{i} \cdot \mathbf{z}_{j}}{\Vert \mathbf{z}_i \Vert_{2} \Vert \mathbf{z}_j \Vert_{2}}$, where $\cdot$ is the dot production of two vectors.
Similar to DAC~\cite{chang2017deep}, we can construct the pseudo-graph $\mathbf{W}$ by setting a large threshold $thres_{1}$:
\begin{equation}
	W_{ij} =
	\left\{
	\begin{aligned}
		&1, \ \  \text{if} \ \ s_{ij} \geq thres_1,\\
		&0, \ \ \text{otherwise.} \\
	\end{aligned}
	\right.
	\label{eq:psedudo_graph}
	\vspace{-0.5mm}
\end{equation}
If the similarity between two samples is larger than the threshold, then we judge that these two samples belong to the same class~($W_{ij}=1$), and the similarity of these samples should be maximized. 
Otherwise~($W_{ij}=0$), the similarity of these samples should be minimized. 
The pseudo-graph supervision can be defined by:\footnote{For the loss function $\ell_g$, there are many choices, such as the contrastive Siamese net loss~\cite{bromley1994signature, luo2018smooth} regularizing the distance between two samples, and the binary cross-entropy loss~\cite{chang2017deep} regularizing the similarity.}
\begin{equation}
	\min_{\theta} L_{PG}(\theta) = \sum_{\mathbf{x}_{i},\mathbf{x}_{j}\in \boldsymbol{\mathcal{X}}} \ell_g(f_{\theta}(\mathbf{x}_{i}),f_{\theta}(\mathbf{x}_{j}); W_{ij} ).
	\label{eq:graph_loss}
	\vspace{-0.5mm}
\end{equation}
%For the loss function $\ell_g$, there are many choices, such as the contrastive Siamese net loss~\cite{bromley1994signature, luo2018smooth} regularizing the distance between two samples, and the binary cross-entropy loss~\cite{chang2017deep} regularizing the similarity.
\par
Please note that there are two differences between our pseudo-graph and that in DAC~\cite{chang2017deep}: 1) Unlike the strong  $\ell_{2}$-norm constrain in DAC, we relax this assumption which only needs to take the output after softmax layer. This relaxation increases the capacity of labeling feature and finally induces a better result in our experiment. 2) Instead of dynamically decreasing threshold in DAC, we only need a fixed threshold of $thres_{1}$. This prevents the training from the disadvantage caused by noisy false positive pairs.

%------------------------------------------------------------------ 
\subsection{Pseudo-label Supervision}
%------------------------------------------------------------------ 

The correlation explored in pseudo-graph is not transitive and limited to pair-wise samples.
Towards this issue, in this subsection, we propose the novel pseudo-label loss and prove its rationality. We first prove the existence of $K$-partition of the pseudo-graph, which could be naturally regarded as pseudo-label. And then we state that this partition would make the optimal solution $\theta^*$ in Eq.~(\ref{eq:graph_loss}) lead to one-hot prediction, which formulates the pseudo-label. Finally, the pseudo-label loss will be introduced to optimize convolutional neural networks.

%\paragraph{Existence of $K$-partition.}
\noindent
\textbf{Existence of $K$-partition.}
The binary relation $W_{ij}$ between samples $\mathbf{x}_i$ and $\mathbf{x}_j$ defined in Eq.~(\ref{eq:graph_loss}) is not transitive: $W_{ij}$ is not deterministic given $W_{ik}$ and $W_{jk}$, and this may lead to unstability in training. Therefore, we introduce Lemma~(\ref{lemma_graph}) to extend it to a stronger relation.
%, by firstly proving the existence of $K$-partitions among $N$ samples~($N\ge K$).
\begin{lemma}\label{lemma_graph}
	For any weighted complete graph $G=(V,E)$ with weight $\omega(e)$ for edge $e$, if $\omega(e_i)\neq \omega(e_j)$ for $\forall i\neq j$, then there exists a threshold $t$ that $G_t=(V, E_t)$ has exactly $K$ partitions, where
	\begin{equation}
		E_t=\{e_i|\omega(e)>t, e_i\in E\}.
	\end{equation}
\end{lemma}
\noindent 
If we take the assumption that $S_{ij}$ is distinctive to each other in similarity graph $\mathbf{S}$, it can be seen as a weighted complete graph under the assumption of Lemma~(\ref{lemma_graph}). Then there exists a threshold $t$ dividing $\boldsymbol{\mathcal{X}}$ into exactly $K$ partitions $\{P^1,P^2, \cdots, P^K\}$. 

\noindent
\textbf{Formulation of the Pseudo-label.}
Let $\mathbf{x}^k$ denote the sample belongs to partition $P^k$, and we can define a transitive relation $\delta$ as:
\begin{equation}
	\delta(\mathbf{x}_i^l,\mathbf{x}_j^k)=
	\left\{
	\begin{aligned}
		&1, \ \  \text{if} \ \ l=k,\\
		&0, \ \ \text{otherwise,} \\
	\end{aligned}
	\right.
\end{equation}
which indicates that pairs with high cosine similarity are guaranteed to be in the same partition. This is to say, as the quality of similarity matrix $\mathbf{S}$ increases during training, this partition gets closer to the ground truth partition, therefore can be regarded as a target to guide and speed up training. Hence, we set the partition $k$ of each $\mathbf{x}$ as its pseudo label.

%The following claim reveals the satisfying relationship between the assigned pseudo label and prediction after the softmax layer in Eq.~(\ref{eq:z_property}):
The following claim reveals the relationship between the assigned pseudo-label and the prediction after softmax:
\begin{claim}\footnote{The proof is presented in supplementary materials.}
	\label{claim_one_hot}
	Let $\theta^*$ denote the optimal solution to Eq.~(\ref{eq:graph_loss}). If $\mathbf{W}$ has $K$ partitions, then the prediction would be one-hot:
	\vspace{-1mm}
	\begin{equation}
		f_{\theta^*}(\mathbf{x}) = (0,\cdots, 0, 1, 0, \cdots,0),\quad \text{for}\; \forall \mathbf{x}.
	\end{equation}
\end{claim}
Hence we can formulate our pseudo-label as:

\vspace{-1mm}
\begin{equation}
	y_{i} = \argmax_{k} \ [f_{\theta}(\mathbf{x}_{i})]_{k},
	\label{eq:pseudo_label}
	\vspace{-1mm}
\end{equation}
where $[\cdot]_{k}$ denotes the $k$-th component of the prediction vector. 
Its corresponding probability of the predicted pseudo-label can be computed by $p_{i} = \max \ [f_{\theta}(\mathbf{x}_{i})]_{k}$.
In practice, $f_{\theta}(\mathbf{x}_{i})$ does not strictly follow the one-hot property, 
since it is difficult to attain the optimal solution for the problem in Eq.~(\ref{eq:graph_loss}) due to the non-convex property.
So we also set a large threshold $thres_2$ for probability $p_{i}$ to select highly-confident pseudo-label for supervision:
\begin{equation}
	V_{i} =
	\left\{
	\begin{aligned}
		&1, \ \  \text{if} \ \ p_{i} \geq thres_2,\\
		&0, \ \ \text{otherwise.} \\
	\end{aligned}
	\right.
	\label{eq:psedudo_label}
\end{equation}
$V_{i}=1$ indicates the predicted pseudo-label is highly-confident, and only
under this situation, will the pseudo-label $y_{i}$ of the $i$-th samples join the network training.  

\noindent
\textbf{Pseudo-label Loss.}
The pseudo-label supervision loss is formulated as:
\begin{equation}
	L_{PL}(\theta) = \sum_{\mathbf{x}_{i} \in \boldsymbol{\mathcal{X}}} V_{i} \cdot \ell_l\left(f_{\theta}(\mathbf{x}_{i}), y_{i} \right).
	\label{eq:label_loss}
\end{equation}
The loss function $\ell_l$ is often defined by the cross-entropy loss.
By combining the supervision of highly-confident pseudo-graph and pseudo-label, we explore the correlation among different samples by minimizing:
\begin{equation}
	L_{CDS} = L_{PG}(\theta) + \alpha L_{PL}(\theta),
	\label{eq:among_corr_loss}
\end{equation}
where $\alpha$ is a balance parameter. Those selected highly-confident information can supervise the training of deep network in a progressive manner.
%---------------------------------------------------------
\subsection{The Local Robustness}
%---------------------------------------------------------
An ideal image representation should be invariant to the geometry transformation, which can be regarded as the local robustness assumption.
Mathematically, given an image sample $\mathbf{x}$ and a geometry transformation $\mathbf{G}$, we denote $\mathbf{x}' = \mathbf{G}\cdot \mathbf{x}$ as the transformed sample, then a good feature extractor $f_{\theta}$ should satisfy that these two samples have the same label and $f_{\theta}( \mathbf{x}) \approx f_{\theta}(\mathbf{x}')$.
Thus we can incorporate the distance between $f_{\theta}(\mathbf{x})$ and $f_{\theta}(\mathbf{x}')$ as a feature invariant loss as:
\vspace{-1mm}
\begin{equation}
	\min_{\theta} \sum_{i=1}^{N} \ell_{r} \left( f_{\theta}(\mathbf{x}_{i}), f_{\theta}(\mathbf{x}_{i}') \right),
	\label{eq:robust1_loss}
\end{equation}
where $\ell_{r}$ is the $\ell_{2}$-norm to measure the distance between predictions of original and transformed samples.
$\mathbf{x}$ and $\mathbf{G}\cdot \mathbf{x}$ generated by the transformation can be regarded as the 'easy' positive pair, which can well stabilize the training and boost the performance.

%It is natural to assume that if two samples are similar to each other in the observation space, they should have the same label, which can be regarded as the local robustness assumption.
%To make the deep unsupervised network more robust, we hope to make full use of this kind of correlation.
%Basically, we add small perturbation or noise on input sample in the observation space, and we hope the prediction of perturbed sample should be consistent with that of the original sample.
% Denote the original sample $\mathbf{x}$ as the parent image, and the perturbed sample $\mathbf{x}'$ as the child image, which 
%is obtained by $\mathbf{x}' = g(\mathbf{x};\xi)$, where $g$ denotes the mapping function with small random perturbation $\xi$.
%Then the local robustness can be achieved by:
%\begin{equation}
%	\min_{\theta} \sum_{i=1}^{N} \ell_{r} \left( f_{\theta}(\mathbf{x}_{i}), f_{\theta}(\mathbf{x}{'}_{i}) \right),
%	\label{eq:robust1_loss}
%\end{equation}
%where $\ell_{r}$ measures the distance between predictions of parent and child samples.
\par
Moreover, please recall that for the original samples, we compute the pseudo-graph and pseudo-label as supervision.
Instead of simply minimizing the distance of predictions, we hope the graph and label information computed based on transformed samples should be consistent with those of original samples.
On the one hand, given an image $\mathbf{x}_{i}$ with highly-confident pseudo-label $y_{i}$, we also force $\mathbf{x}_{i}'$ has same pseudo-label.
On the other hand, we also investigate the correlation among the transformed samples $\mathbf{x}'$ with the highly-confident pseudo-graph $\mathbf{W}$ computed on the original samples $\mathbf{x}_{i}$, which is beneficial to increase the network robustness.
The loss function to achieve above targets can be formulated as:
\begin{align}
	L_{LR}
	& \! = \!  \!  \sum_{\mathbf{x}'_{i},\mathbf{x}'_{j} \in \boldsymbol{\mathcal{X}}'}  \!  \!  \!  \ell_g(f_{\theta}(\mathbf{x}'_{i}),f_{\theta}(\mathbf{x}'_{j}); W_{ij} )  \! + \! 
	\alpha \! \!   \!  \sum_{\mathbf{x}'_{i} \in \boldsymbol{\mathcal{X}}'}  \! \!   \! 
	V_{i} \!  \cdot \!  \ell_l\left(f_{\theta}(\mathbf{x}'_{i}), y_{i} \right) \notag \\
	&  \! = L'_{PG}(\theta) + \alpha L'_{PL}(\theta),
	\label{eq:robust_loss}
\end{align}
where $\boldsymbol{\mathcal{X}}' = \{ \mathbf{x}'_{i} \}_{i=1}^{N}$ is the transformed data set, $\mathbf{W}$ and $\mathbf{V}$ are same to those of original set in Eqs.~(\ref{eq:psedudo_graph}) and~(\ref{eq:psedudo_label}).
\par
The deep unsupervised learning can benefit a lot from the above strategy.
As we set high confidence for the construction of pseudo-graph and pseudo-label, it can be regarded as the easy sample, which will contribute little to the parameter learning~\cite{8578392}.
By adding small perturbation, the prediction of transformed sample will not be easy as that of original sample, which will contribute a lot in return.

%------------------------------------------------
%------------------------------------------------
\subsection{Triplet Mutual Information}\label{sec:set_MI}
%------------------------------------------------
%Besides mining correlation among samples and local robustness, 
In this section, we explore the correlation between deep and shallow layer representations of each instance and propose a novel loss, named triplet mutual information loss, to make full use of the feature correspondence information. Firstly, we introduce the mutual information loss which is proposed in~\cite{nowozin2016f,hjelm2018learning} and analyze its limitation. 
Next, the concept of triplet correlations is described.
Finally, we propose the triplet mutual information loss that enables convolutional neural networks to learn discriminative features. 

%
%we make full use of the correlation between deep and shallow layer representations of each instance .
%whose mutual information~(MI) should be maximized.
The mutual information~(MI) between deep and shallow layer features of the same sample should be maximized, which guarantees the consistency of representation.
Similar to~\cite{nowozin2016f}, we also convert the MI of two random variables~($D$ and $S$) to the Jensen-Shannon divergence~(JSD) between samples coming from the joint distribution $\mathbb{J}$ and their product of marginals $\mathbb{M}$.
Correspondingly, features of different layers should follow the joint distribution only when they are features of the same sample, otherwise, they follow the marginal product distribution.
So JSD version MI is defined as:
\begin{equation}
	\mathcal{MI}^{(JSD)}(D, S) = \mathbb{E_{J}}[-{\rm sp}(-T(d, s))] - \mathbb{E_M}[{\rm sp}(T(d, s))],
	\label{eq:MI_JSD1}
\end{equation}
where $d$ corresponds to the deep layer features, $s$ corresponds to the shallow layer features,  
$T$ is a discriminator trained to distinguish whether $d$ and $s$ are sampled from the joint distribution or not,
and ${\rm sp}(z) = \log(1+e^{z})$ is the softplus function. 
For discriminator implementation, \cite{hjelm2018learning} shows that incorporating knowledge about locality in the input can improve the representations' quality.
%The implementation of the discriminator varies and~\cite{hjelm2018learning} shows that incorporating knowledge about locality in the input can improve the representations' quality, hence we adopt the local strategy, maximizing the average MI between high-level representation and local patches of the image by concatenating the replicate of vectors with map at each coordinate as the input of the discriminator. 
\par
%While improving the stability and consistence in unsupervised training, we also want our representation to be more discriminative to suit our downstream tasks. 
Please note that currently, we do not incorporate any class information.
For two different samples $\mathbf{x}_{1}$ and $\mathbf{x}_{2}$, the mutual information between $\mathbf{x}_{1}$'s shallow-layer representation and $\mathbf{x}_{2}$'s deep-layer representation will be minimized even if they belong to the same class, which is not reasonable.
%
%Noticing that currently we are encouraging representations of samples in the same class to appear differently -- by minimizing the mutual information between x's low-level representation and y's high-level representation, where x and y are samples from the same class -- which makes no sense, 
So we consider fixing this issue by introducing the mutual information loss of positive pairs. 
As shown in the bottom right of Figure~\ref{fig:pipline},
with the generated pseudo-graph $\mathbf{W}$ described in Section~\ref{subsection:pg}, we select positive pairs and negative pairs with the same anchor to construct triplet correlations.
Analogous to supervised learning, this approach lifts the instance-level mutual information supervision to triplet-level supervision.

%Specifically, we set two thresholds to the CosSimilarity value to fetch positive pairs and negative pairs, and the thresholds need to be adjusted carefully to balance the trade-off between confidence and training difficulty in unsupervised learning task. (As illustrated in XXX, easy examples make little contribution in network training; while XXX shows that noise leads to even worse convergency.) 
%In the meantime, we removed those selected positive pairs from the negative concatenation list.
%With positive pairs and corresponding negative pairs, we construct triplet correlation.
\par
Then we show how this approach is theoretically formulated by extending Eq.~(\ref{eq:MI_JSD1}).
%Then we extend the formulation in Eq.~(\ref{eq:MI_JSD1}) to triplet mutual information loss.
We set the samples of random variable $D$ and $S$ to be sets, instead of instances. Denote the deep layer feature of sample $j$ belongs to class $i$ as $d^i_j$ and its shallow layer feature as $s_j^i$, then $D^i=\{d^i_1, d_2^i, \cdots, d_n^i\}$ and $S^i=\{s^i_1, s^i_2, \cdots, s^i_n\}$ are feature sets of class $i$.
Variables $\mathbb{D}$ and $\mathbb{S}$ are defined by $\mathbb{D}= \{D^1, D^2, \cdots, D^K\}$ and $\mathbb{S}= \{S^1, S^2, \cdots, S^K\}$, respectively.
Then we can get the following extension of Eq.~(\ref{eq:MI_JSD1}):
%following formulation:
%\begin{align}
%\mathcal{MI}^{(JSD)}_{set}(D, S) = & \mathbb{E_{(D, S)\sim J}}[-{\rm sp}(-T(d, s))] \label{eq:MI_set}  \\ \notag
%& - \mathbb{E_{D\times S\sim M}}[{\rm sp}(T(d, s))],
%\end{align}
\begin{align}
	L_{MI} \! =  \!  -\mathcal{MI}^{(JSD)}_{set} (D,  S & )   \! =  -\left( \mathbb{E_{(D, S)= J}}[-{\rm sp}(-T(d, s))] \label{eq:MI_set}  \right. \notag \\ 
	&  \left. - \mathbb{E_{D\times S= M}}[{\rm sp}(T(d, s))] \right),
\end{align}
where we investigate the mutual information based on class-related feature sets.
In this case, besides considering the features of same sample,
we also maximize the mutual information between different layers' features for samples belongs to the same class.
The overview of triplet mutual information loss is shown in the bottom right of Figure~\ref{fig:pipline}.
%That is to say, the samples of variable $\mathbb{D}$ and $\mathbb{S}$ are sets, and maximizing the MI between them means that we are building the bridge between feature sets $D^i$ and $S^i$ instead of between samples $d_m$ and $s_m$ of the same instance, which can encourage $s^i_m$ and $d^i_n$ having stronger causality for $\forall m, n$  as long as they belong to the same class $i$. 
Specifically, we compute the loss function in Eq.~(\ref{eq:MI_set}) by pair-wise sampling. 
For each sample, we construct the positive pairs and negative pairs based on the pseudo-graph $\textbf{W}$ to compute the triplet mutual information loss, which is very helpful to learn more discriminative representations.

%as we sample pairs $s^i_m$ and $d^j_n$ and tag them positive concatenation if $i=j$ otherwise negative, and whether $i=j$ will be determined by the pseudo-graph $\textbf{W}$. Through this approach, we encourage the encoded representations of $d^i_j$ and $d^i_k$ becoming similar, which means discriminativeness for our task. This action is under the assumption that the feature vector is of limited capacity, so the encoder is encouraged to pick the general character that is shared within the class. Otherwise, if the encoder focuses on irrelevant informations such as noises or transformations, it does not increase the MI of positive pairs. This strategy can significantly boost the quality of representation.

%------------------------------------------------------------------
\renewcommand{\algorithmicrequire}{\textbf{Input:}}
\renewcommand{\algorithmicensure}{\textbf{Output:}}
\begin{algorithm}[!tp]
	\caption{Deep Comprehensive Correlation Mining}
	\label{alg:DCCM}
	\begin{algorithmic}[1]
		\REQUIRE  Unlabeled dataset $\boldsymbol{\mathcal{X}} = \{ \mathbf{x}_{i} \}_{i=1}^{N}$, $thres_{1}$, $thres_2$. \\
		\STATE Initialize the network parameter $\theta$ randomly;
		\STATE \textbf{for} $t$ in $[1, num\_epoches ]$ \textbf{do}	\\	
		\STATE \quad \textbf{for} each minibatch $\boldsymbol{\mathcal{X_{B}}}$  \textbf{do} \\
		\STATE \qquad Compute the prediction feature $f(\mathbf{x}_{i})$ for each \\ \qquad sample $\mathbf{x}_{i}$  in the minibatch set $\boldsymbol{\mathcal{X_{B}}}$;
		\STATE \qquad Compute the similarity $s_{ij}$,   pseudo-graph $\mathbf{W}$  \\ \qquad and pseudo-label based on Eqs.~(\ref{eq:psedudo_graph}),~(\ref{eq:pseudo_label}) and~(\ref{eq:psedudo_label});
		\STATE \qquad Select positive and negative pairs based on $\mathbf{W}$;
		\STATE \qquad Compute the DCCM loss by Eq.~{(\ref{eq:final_obj})};
		\STATE \qquad Update $\theta$ using optimizers;
		\STATE \quad \textbf{end for}		
		\STATE \textbf{end for}
		\ENSURE  Compute the cluster label by Eq.~(\ref{eq:pseudo_label}).
	\end{algorithmic}
\end{algorithm}
%------------------------------------------------------------------

%------------------------------------------------
\subsection{The Unified Model and Optimization}\label{unified}
%------------------------------------------------
By combining the investigations of these three aspects in above subsections and jointly train the network, we come up with our deep comprehensive correlation mining for unsupervised learning and clustering. The final objective function of DCCM can be formulated as:

%%---------------------------
%\begin{equation}
%L_{DCCM} = L_{CDS} + \beta L_{LR} + \gamma L_{FC},
%\label{eq:final_obj}
%\end{equation}
%%---------------------------
%---------------------------
\begin{equation}
	\min_{\theta} L_{DCCM} = \widehat{L_{PG}} + \alpha \widehat{L_{PL}} + \beta L_{MI},
	\label{eq:final_obj}
\end{equation}
%---------------------------
where $\alpha$ and $\beta$ are constants to balance the contributions of different terms, $\widehat{L_{PG}} = L_{PG} + L'_{PG}$ is the overall pseudo-graph loss, and $\widehat{L_{PL}} = L_{PL} + L'_{PL}$ is the overall pseudo-label loss.
The framework of DCCM is presented in Figure~\ref{fig:pipline}.
Based on the ideally one-hot prediction feature, we compute the highly-confident pseudo-graph and pseudo-label to guide the feature learning of both original and transformed samples, investigating both correlations among different samples and local robustness for small perturbation.
In the meantime, to investigate feature correspondence for discriminative feature learning, the pseudo-graph is also utilized to select highly-confident positive and negative pairs for triplet mutual information optimization.
\par
Our proposed method can be trained in a minibatch based end-to-end way, which can be optimized efficiently. 
After the training, the predicted feature is ideally one-hot.
The predicted cluster label for sample $\textbf{x}_{i}$ is exactly same to the pseudo-label $y_{i}$, which is easily computed by Eq.~(\ref{eq:pseudo_label}).
We summarize the overall training process in Algorithm~\ref{alg:DCCM}.
%------------------------------------------------------------------------
\begin{table}[!t]	
	%\footnotesize
	\small
	\centering
	\renewcommand{\arraystretch}{0.92}
	\begin{centering}
		\begin{threeparttable}[]
			\caption{Statistics of different datasets.}\label{tab:datasets}
			\tabcolsep=1pt
			\begin{minipage}{12cm}
				\begin{tabular}{@{}lcccc}
					\toprule[0.45pt] %\toprule define the top line width
					Dataset         &Train Images &Test Images &Clusters &Image size  \\ \hline
					CIFAR-10               &$50,000$ &$10,000$  &10 & $32\times 32 \times 3$ \\
					CIFAR-100    &$50,000$ &$10,000$   &20/100 & $32\times 32 \times 3$\\
					STL-10    &$13,000$ &--   &10 & $96\times 96 \times 3$\\
					ImageNet-10               &$13,000$ &--   &10 & $96\times 96 \times 3$ \\
					ImageNet-dog-15           &$19,500$ &--   &15 & $96\times 96 \times 3$ \\
					Tiny-ImageNet           &$100,000$  &--  &200 & $64\times 64 \times 3$\\  \bottomrule[0.45pt]
				\end{tabular}
			\end{minipage}
		\end{threeparttable}
	\end{centering}
\vspace{-4mm}
\end{table}
\begin{table*}[!t]
	\renewcommand\arraystretch{1.1}
	\footnotesize
	\caption{Clustering performance of different methods on six challenging datasets. The best results are highlighted in \textbf{bold}.}
	\label{tab:clustering_res}
	\setlength{\tabcolsep}{3.2pt}
	\begin{tabular}{|c|ccc|ccc|ccc|ccc|ccc|ccc|}
		
		\hline
		Datasets                       & \multicolumn{3}{c|}{CIFAR-10}                       & \multicolumn{3}{c|}{CIFAR-100}                      & \multicolumn{3}{c|}{STL-10}       & \multicolumn{3}{c|}{ImageNet-10} & \multicolumn{3}{c|}{Imagenet-dog-15}                 & \multicolumn{3}{c|}{Tiny-ImageNet}                  \\ \hline
		Methods\textbackslash{}Metrics & NMI      & ACC    & ARI & NMI      & ACC     & ARI & NMI    & ACC        & ARI & NMI                   & ACC                   & ARI & NMI                   & ACC                   & ARI & NMI                   & ACC                   & ARI \\ \hline
		K-means & 0.087 & 0.229 & 0.049 & 0.084 & 0.130 & 0.028 & 0.125 & 0.192 & 0.061 & 0.119  & 0.241  & 0.057  & 0.055  & 0.105  & 0.020 & 0.065 & 0.025   &  0.005 \\ \hline
		SC~\cite{zelnik2005self}      & 0.103 & 0.247 & 0.085 & 0.090 & 0.136 & 0.022 & 0.098 & 0.159 & 0.048 & 0.151  & 0.274  & 0.076  & 0.038  & 0.111  & 0.013 & 0.063  & 0.022   & 0.004  \\ \hline
		AC~\cite{gowda1978agglomerative}      & 0.105 & 0.228 & 0.065 & 0.098 & 0.138 & 0.034 & 0.239 & 0.332 & 0.140 & 0.138  & 0.242  & 0.067  & 0.037  & 0.139  & 0.021 & 0.069  & 0.027   & 0.005 \\ \hline
		NMF~\cite{cai2009locality}     & 0.081 & 0.190 & 0.034 & 0.079 & 0.118 & 0.026 & 0.096 & 0.180 & 0.046 & 0.132  & 0.230  & 0.065  & 0.044  & 0.118  & 0.016 & 0.072 & 0.029  & 0.005  \\ \hline
		AE~\cite{bengio2007greedy}      & 0.239 & 0.314 & 0.169 & 0.100 & 0.165 & 0.048 & 0.250 & 0.303 & 0.161 & 0.210  & 0.317  & 0.152  & 0.104  & 0.185  & 0.073 & 0.131 & 0.041  & 0.007 \\ \hline
		DAE~\cite{vincent2010stacked}     & 0.251 & 0.297 & 0.163 & 0.111 & 0.151 & 0.046 & 0.224 & 0.302 & 0.152 & 0.206  & 0.304  & 0.138  & 0.104  & 0.190  & 0.078 & 0.127 & 0.039   & 0.007  \\ \hline
		GAN~\cite{radford2015unsupervised}     & 0.265 & 0.315 & 0.176 & 0.120 & 0.151 & 0.045 & 0.210 & 0.298 & 0.139 & 0.225  & 0.346  & 0.157  & 0.121  & 0.174  & 0.078 & 0.135 & 0.041   & 0.007  \\ \hline
		DeCNN~\cite{zeiler2010deconvolutional}   & 0.240 & 0.282 & 0.174 & 0.092 & 0.133 & 0.038 & 0.227 & 0.299 & 0.162 & 0.186  & 0.313  & 0.142  & 0.098  & 0.175  & 0.073 & 0.111 & 0.035   & 0.006  \\ \hline
		VAE~\cite{kingma2013auto}     & 0.245 & 0.291 & 0.167 & 0.108 & 0.152 & 0.040 & 0.200 & 0.282 & 0.146 & 0.193  & 0.334  & 0.168  & 0.107  & 0.179  & 0.079 & 0.113 & 0.036  & 0.006 \\ \hline
		JULE~\cite{yang2016joint}    & 0.192 & 0.272 & 0.138 & 0.103 & 0.137 & 0.033 & 0.182 & 0.277 & 0.164 & 0.175  & 0.300  & 0.138  & 0.054  & 0.138  & 0.028 & 0.102 & 0.033  & 0.006 \\ \hline
		DEC~\cite{2016ICMLDEC}     & 0.257 & 0.301 & 0.161 & 0.136 & 0.185 & 0.050 & 0.276 & 0.359 & 0.186 & 0.282  & 0.381  & 0.203  & 0.122  & 0.195  & 0.079 & 0.115  & 0.037   & 0.007  \\ \hline
		DAC~\cite{chang2017deep}     & 0.396 & 0.522 & 0.306 & 0.185 & 0.238 & 0.088 & 0.366 & 0.470 & 0.257 & 0.394  & 0.527  & 0.302  & 0.219  & 0.275  & 0.111 & 0.190  & 0.066  & 0.017 \\ \hline
		DCCM~(ours)    & \textbf{0.496}       &   \textbf{0.623}         & \textbf{0.408}    & \textbf{0.285}     & \textbf{0.327}     & \textbf{0.173}        &  \textbf{0.376}          & \textbf{0.482} &  \textbf{0.262}      &     \textbf{0.608}       & \textbf{0.710} &   \textbf{0.555}     &      \textbf{0.321}      &  \textbf{0.383} &  \textbf{0.182}    &    \textbf{0.224}     &  \textbf{0.108}          &  \textbf{0.038}   \\ \hline
	\end{tabular}
\vspace{-6mm}
\end{table*}
%------------------------------------------------
%------------------------------------------------
\section{Experiments}
%------------------------------------------------
We distribute our experiments into a few sections. We first examine the effectiveness of DCCM by comparing it against other state-of-the-art algorithms. After that, we conduct more ablation studies by controlling several influence factors. Finally, we do a series of analysis experiments to verify the effectiveness of the unified model training framework. Next, we introduce the experimental setting.
%In this section, we conduct extensive experiments to analyze our deep comprehensive correlation mining on two different tasks, including the deep clustering task based on the predicted labels, and the transfer learning classification task based on the learned feature representations.
%------------------------------------------------
%\subsection{Experimental Settings}
%------------------------------------------------

%------------------------------------------------

\noindent
\textbf{Datasets.}
We select six challenging image datasets for deep unsupervised learning and clustering, including the CIFAR-10~\cite{cifar10}, CIFAR-100~\cite{cifar10}, STL-10~\cite{coates2011analysis}, Imagenet-10, and ImageNet-dog-15, and Tiny-ImageNet~\cite{deng2009imagenet} datasets.
We summarize the statistics of these datasets in Table~\ref{tab:datasets}.
\par
For the clustering task, we adopt the same setting as that in~\cite{chang2017deep}, where the training and validation images of each dataset are jointly utilized, and the $20$ superclasses are considered for the CIFAR-100 dataset in experiments.
ImageNet-10 and ImageNet-dog-15 used in our experiments are same  as~\cite{chang2017deep}, where they randomly choose $10$ subjects and $15$ kinds of dog images from the ImageNet dataset, and resize these
images to $96 \times 96 \times 3$. 
As for the Tiny-ImageNet dataset, a reduced version of the ImageNet dataset~\cite{deng2009imagenet}, it totally contains $200$ classes of $110,000$ images, which is a very challenging dataset for clustering.
\par
For the transfer learning classification task, we adopt the similar setting as that in~\cite{hjelm2018learning}, where we mainly consider the CIFAR-10, CIFAR-100 of $100$ classes. Training and testing samples are separated.
%\vspace{-2mm}
%------------------------------------------------

\noindent
\textbf{Evaluation Metrics.}
To evaluate the performance of clustering, we adopt three commonly used metrics including normalized mutual information~(NMI), accuracy~(ACC), adjusted rand index~(ARI).
These three metrics favour different properties in clustering task. For details, please refer to the appendix.
For all three metrics, the higher value indicates the better performance.
\par
To evaluate the quality of feature representation, we adopt the non-linear classification task which is the same as that in~\cite{hjelm2018learning}.
%Specifically, for the linear classification, features of training set are used to train a support vector machine~(SVM) after which we test the linear separability on the validation set.
Specifically, 
after the training of DCCM, we fix the parameter of deep neural network and train a multilayer perception network with a single hidden layer~($200$ units) on top of the last convolutional layer and fully-connected layer features separately in a supervised way. 

%------------------------------------------------
%\subsubsection{Compared Methods}
%------------------------------------------------

%\quad \ We compare our DCCM with the state-of-the-art methods of different tasks separately.
\par

\par
%For classification task, we compare DCCM against several unsupervised feature learning methods, including variational auto-encoder~(VAE)~\cite{kingma2013auto}, adversarial auto-encoder~(AAE)~\cite{makhzani2015adversarial}, BiGAN~\cite{donahue2016adversarial}, noise as targets~(NAT)~\cite{bojanowski2017unsupervised}, and deep infomax~(DIM)~\cite{hjelm2018learning}.

%\vspace{-2mm}
%------------------------------------------------
%------------------------------------------------------------------
\begin{figure*}[!ht]
	\centering 
	\subfigure[Initial stage of DCCM]{ 
		\label{fig:ori} %% label for second subfigure 
		\includegraphics[width=1.78in]{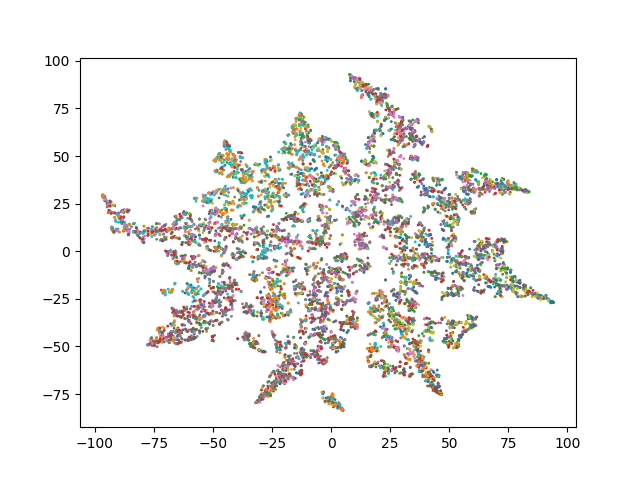}} 
	\hspace{-6mm} 
	\subfigure[Middle stage of DCCM]{ 
		{\label{fig:middle}} %% label for first subfigure 
		\includegraphics[width=1.78in]{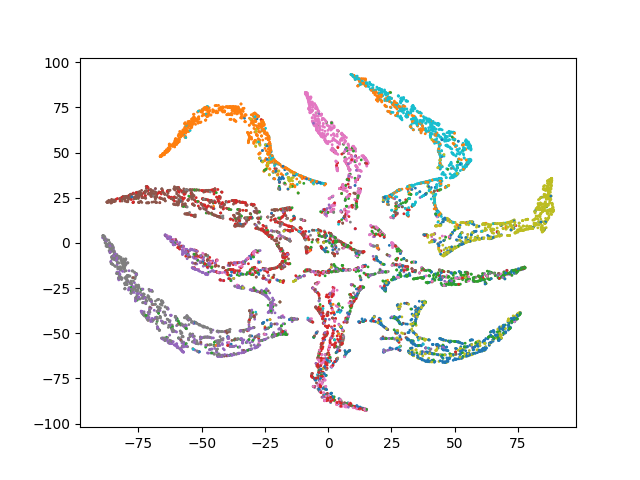}} 
	\hspace{-6mm}
	\subfigure[Final stage of DCCM]{ 
		{\label{fig:final}} %% label for first subfigure 
		\includegraphics[width=1.78in]{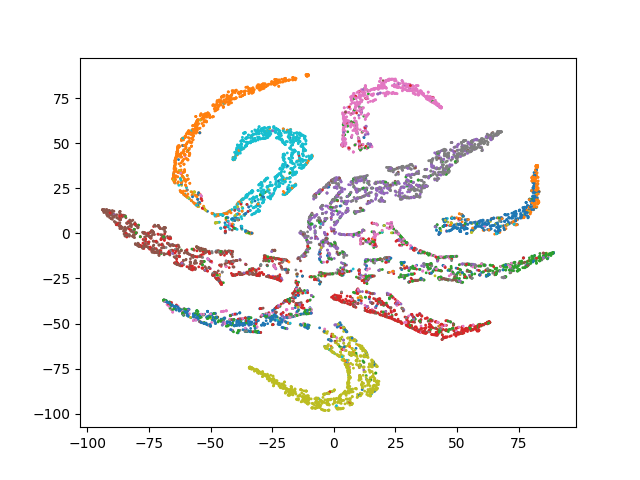}} 
	\hspace{-6mm}
	\subfigure[Final stage of DAC]{ 
		{\label{fig:DAC}} %% label for first subfigure 
		\includegraphics[width=1.78in]{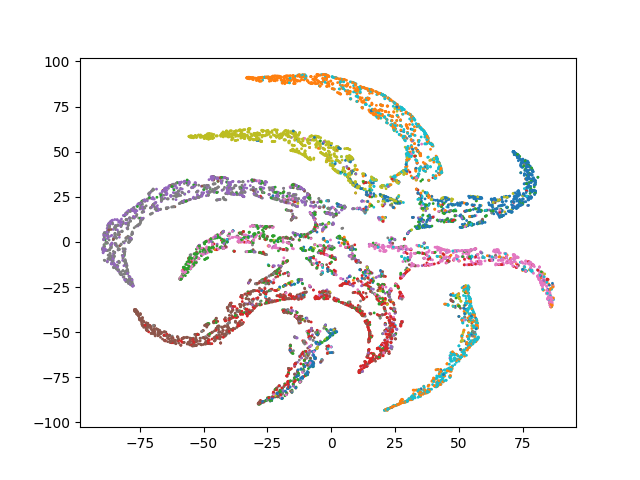}} 
	\caption{Visualizations of embeddings for different stages of DCCM and DAC on the CIFAR-10 dataset. Different colors denote various clusters. From (a) to (c), with the increasing of epochs, DCCM tends to progressively learn more discriminative features. Based on (c) and (d), features of DCCM are more discriminative than that of DAC.} 
	\label{fig:VIS} %% label for entire figure 
	\vspace{-4mm}
\end{figure*}
%------------------------------------------------------------------
\noindent
\textbf{Implementation Details.}
The network architecture used in our framework is a shallow version of the AlexNet~(details for different datasets are described in the supplementary materials). 
Similar to~\cite{chang2017deep}, we adopt the RMSprop optimizer with $lr=1e^{-4}$.
For hyper-parameters, we set $\alpha=5$ and $\beta=0.1$ for all datasets, which are relatively stable within a certain range. 
The thresholds to construct highly-confident pseudo-graph and select highly-confident pseudo-label are set to $0.95$ and $0.9$, respectively.
The small perturbations used in the experiments include rotation, shift, rescale, etc.
For discriminator of mutual information estimation, we adopt the network with three $1\times 1$ convolutional layers, which is same to~\cite{hjelm2018learning}.
We use pytorch~\cite{paszke2017automatic} to implement our approach.
%\par
%%------------------------------------------------
%\begin{table}[!t]
%	\renewcommand\arraystretch{1.2}
%	\centering
%	\caption{Linear classification accuracy~(top 1) results of different deep unsupervised feature learning methods on the CIFAR-10 dataset. The best result is highlighted in \textbf{bold}.}
%	\label{tab:linear_classify_res}
%	\setlength{\tabcolsep}{10pt}
%	\begin{tabular}{|c|cc|}
%		
%		\hline
%		Methods\textbackslash{}Features & SVM(Conv)         & SVM(Y(64))    \\ \hline
%		VAE       &0.538 &0.396    \\ \hline
%		AAE       &0.552 &0.378    \\ \hline
%		BiGAN     &0.564 &0.449     \\ \hline
%		NAT       &0.486 &0.396   \\ \hline
%		DIM       &0.633 &0.496    \\ \hline
%		DCCM~(ours)    & \textbf{ }       & \textbf{}    \\ \hline
%	\end{tabular}
%\end{table}
%%------------------------------------------------
%------------------------------------------------------------------
\begin{figure}[t]
	\centering 
	\includegraphics[width=\linewidth]{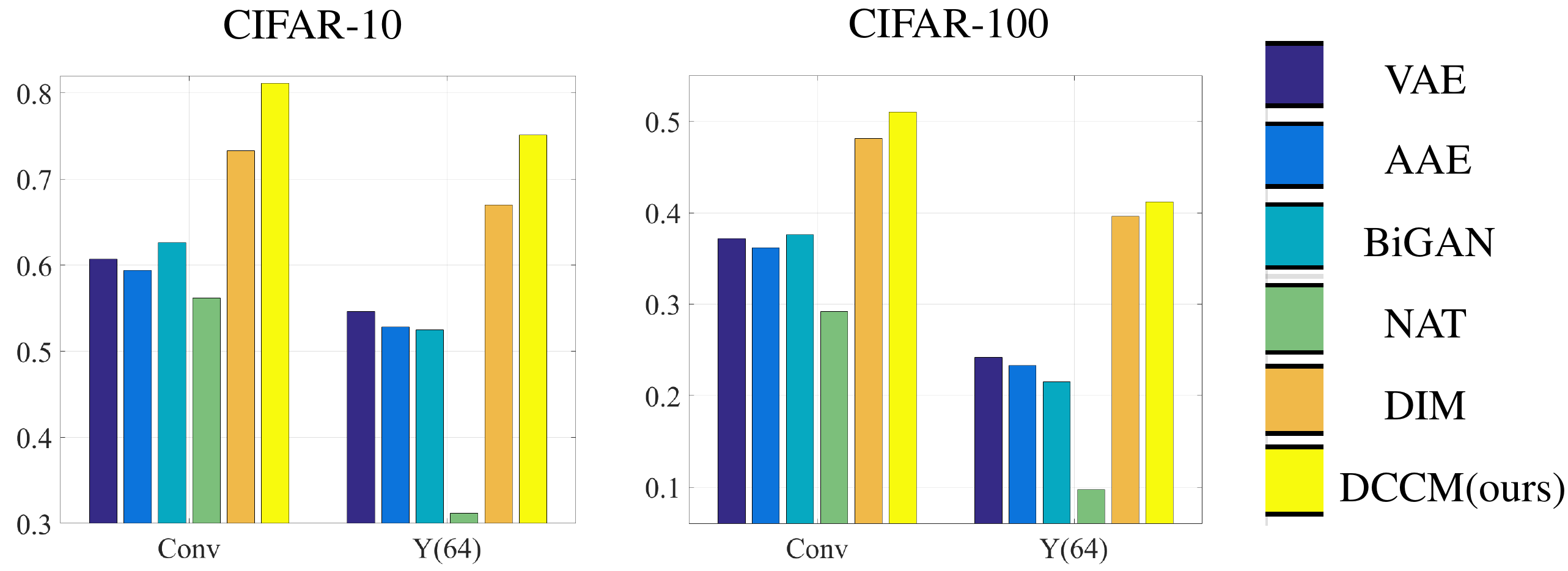}
	\caption{Non-linear classification accuracy~(top 1) results of different deep unsupervised feature learning methods on two datasets. 'Conv' denotes the features after the last convolutional layer, and 'Y($64$)' denotes the $64$-dimensional feature of fully-connected layer.
	} 
	\label{fig:class_acc}
	\vspace{-4mm}
\end{figure}
%------------------------------------------------------------------
%------------------------------------------------
\subsection{Main Results}
%In Table~\ref{tab:clustering_res}, we compare the performance of our method with other state-of-the-art clustering methods on the image clustering task.
We first compare the DCCM with other state-of-the-art clustering methods on the clustering task. 
The results are shown in the Table~\ref{tab:clustering_res}.
Most results of other methods are directly copied from DAC~\cite{chang2017deep}.
DCCM significantly surpasses other methods by a large margin on these benchmarks under all three evaluation metrics.
Concretely, the improvement of DCCM is very significant even compared with the state-of-the-art method DAC~\cite{chang2017deep}.
Take the clustering ACC for example, our result $0.623$ is $10.1\%$ higher than the performance $0.522$ of DAC~\cite{chang2017deep} on the CIFAR-10 dataset.
On the CIFAR-100 dataset, the gain of DCCM is $8.9\%$ over DAC~\cite{chang2017deep}.

Figure~\ref{fig:VIS} visualizes feature embeddings of the DCCM and DAC on CIFAR-10  using t-SNE~\cite{maaten2008visualizing}.
We can see that compared with DAC, DCCM exhibits more discriminative feature representation.
Above results can sufficiently verify the effectiveness and superiority of our proposed DCCM.
\par
%Based on the results, we also have two other observations.
%Due to the powerful representation ability of deep learning, deep clustering methods achieves much better results than traditional methods.
%Secondly, DEC and DAC show better results than other deep feature learning methods, such as AE, GAN, and VAE.
%The reason is that the guidance of clustering loss can lead to much discriminative features as well as better performance.
%
%Besides, with the increasing of number of clusters, more uncertainties are brought in, and the clustering performance tends to decrease.
%Fortunately, even with large number of clusters, our method still achieves significant improvement, which also verifies the robustness of DCCM.

\par
To further evaluate the quality of feature representations, we adopt the classification task and compare DCCM with other deep unsupervised feature learning methods.
We compare DCCM against several unsupervised feature learning methods, including variational AE~(VAE)~\cite{kingma2013auto}, adversarial AE~(AAE)~\cite{makhzani2015adversarial}, BiGAN~\cite{donahue2016adversarial}, noise as targets~(NAT)~\cite{bojanowski2017unsupervised}, and deep infomax~(DIM)~\cite{hjelm2018learning}.
The top 1 non-linear classification accuracy comparison is presented in Figure~\ref{fig:class_acc}.
We can also observe that DCCM achieves much better results than other methods on CIFAR-10 and CIFAR-100 datasets.
Especially on the CIFAR-10 dataset, our results on both convolutional and fully-connected layer features are more than $8\%$ higher than these of the second best method DIM.
Since we incorporate the graph-based class information and transform the instance-level mutual information into the triplet-level, our method can learn much more discriminative features, which accounts for the obvious improvement.
\par
We also compare with several state-of-the-art methods under the same architecture and analyze the influence of various sampling strategy in the supplementary materials.
%
%Even on the very challenging Tiny-ImageNet dataset, DCCM also achieves much better results than DAC.
%
%
%DCCM gains better discriminative feature representation.
%
%The feature embeddings of DCCM exhibits better discriminative feature representation, which suggests that comprehensive correlation mining is 
%In Figure~\ref{fig:VIS}, we visualize the embeddings of the initial and the final stages on the CIFAR-10 dataset using t-SNE~\cite{maaten2008visualizing}. We can see that after training, the representations are very discriminative and easy to classify for different classes.

%------------------------------------------------
%\begin{table}[!t]
%	\renewcommand\arraystretch{1.3}
%	\caption{Ablation study of DCCM on the CIFAR-10 dataset. }
%	\label{tab:ablation}
%	\setlength{\tabcolsep}{2pt}
%	\begin{tabular}{|c|c|ccccc|}

%		\hline    \multirow{2}{*}{}    &    \multirow{2}{*}{Methods}         & \multicolumn{5}{c|}{Metrics}                                      \\ \cline{3-7} 
%		&   & NMI      & ACC    & ARI & Conv      & Y    \\ \hline
%		M1	&	${L_{PG}}$    & 0.304 & 0.405 & 0.232 & 0.742 & 0.677   \\ \hline
%		M2	& $\widehat{L_{PG}}$    & 0.412 & 0.512 & 0.323 & 0.781 & 0.715   \\ \hline
%		M3	& $\widehat{L_{PG}}+\widehat{L_{PL}}$     & 0.448 & 0.583 & 0.358 & 0.804 & 0.740  \\ \hline
%		M4	& $\widehat{L_{PG}}+\widehat{L_{PL}}+L_{MI}$   & \textbf{0.496}       &   \textbf{0.623}         & \textbf{0.408}    & \textbf{0.811}     & \textbf{0.751}    \\ \hline
%	\end{tabular}
%	\vspace{-2mm}
%\end{table}
%------------------------------------------------
\subsection{Correlation Analysis}
%In Table~\ref{tab:ablation}, we present the results of DCCM and its ablated versions.
We analyze the effectiveness of various correlations from three aspects: Local Robustness, Pseudo-label and Triplet Mutual Information in this section. The results are shown in Table~\ref{tab:ablation}.
%Based on the ablation study, we analyze the effectiveness of various correlations mined in this paper.
%We can see that both the overall pseudo-label and extended feature correspondence can contribute a lot to the baseline with only overall pseudo-graph loss, which can well verify the necessity to investigate comprehensive correlations.
%------------------------------------------------
\begin{table}[!t]
	\centering
	\renewcommand\arraystretch{1.3}
	\caption{Ablation study of DCCM on the CIFAR-10 dataset. LR, PL, and MI corresponds to local robustness, pseudo-label, and mutual information, respectively.}
	\label{tab:ablation}
	\setlength{\tabcolsep}{1.8pt}
	\small
	\begin{tabular}{|c|c|ccc|ccc|}
		
		\hline    \multirow{2}{*}{}    &    \multirow{2}{*}{Methods}      & \multicolumn{3}{c|}{Correlations}    & \multicolumn{3}{c|}{Metrics}                                      \\ \cline{3-8} 
		&   & LR & PL  & MI & NMI      & ACC    & ARI   \\ \hline
		M1	&	${L_{PG}}$  &  &    &   & 0.304 & 0.405 & 0.232    \\ \hline
		M2	& $\widehat{L_{PG}}$  & $\checkmark$ &   &   & 0.412 & 0.512 & 0.323    \\ \hline
		M3	& $\widehat{L_{PG}}+\widehat{L_{PL}}$  & $\checkmark$ & $\checkmark$  &    & 0.448 & 0.583 & 0.358   \\ \hline
		M4	& $\widehat{L_{PG}}+\widehat{L_{PL}}+L_{MI}$ & $\checkmark$ & $\checkmark$  & $\checkmark $  & \textbf{0.496}       &   \textbf{0.623}         & \textbf{0.408}       \\ \hline
	\end{tabular}
\vspace{-5mm}
\end{table}
%------------------------------------------------

\noindent
\textbf{Local Robustness Influence.} 
The only difference between methods M2 and M1 lies in whether to use the local robustness mechanism or not. We can see that M2 significantly surpasses the M1, which demonstrates the robustness and effectiveness of local robustness. 
Because we set high threshold to select positive pairs, without transformation, these easy pairs have limited contribution to parameter learning. 
With the local robustness loss, we construct many hard sample pairs to benefit the network training. So it significantly boosts the performance.

\noindent
\textbf{Effectiveness of Pseudo-label.}
With the help of pseudo-label, M3~(with both pseudo-graph and pseudo-label) achieves much better results than M2~(with only pseudo-graph) under all metrics.
Specifically, there is a $7.1\%$ improvement on clustering ACC.
The reason is that pseudo-label can make full use of the category information behind the feature distribution, which can benefit the clustering.

\noindent
\textbf{Triplet Mutual Information Analysis.} 
Comparing the results of M4 and M3, we can see that the triplet mutual information can further improve the clustering ACC by $4.0\%$.
As we analyzed in Section~\ref{sec:set_MI}, with the help of pseudo-graph, triplet mutual information can not only make use of the features correspondence of the same sample, but also introduce discriminative property by constructing positive and negative pairs. So it can further improve the result.

%\subsection{DCCM Framework Analysis}
\subsection{Overall Study of DCCM}
In this section, we conducted experiments on CIFAR-10~\cite{cifar10} to investigate the behavior of deep comprehensive correlations mining. The model is trained with the unified model optimization which is introduced in Section~\ref{unified}.
%------------------------------------------------

\begin{figure}[t]
	\centering 
	\includegraphics[width=0.8\linewidth]{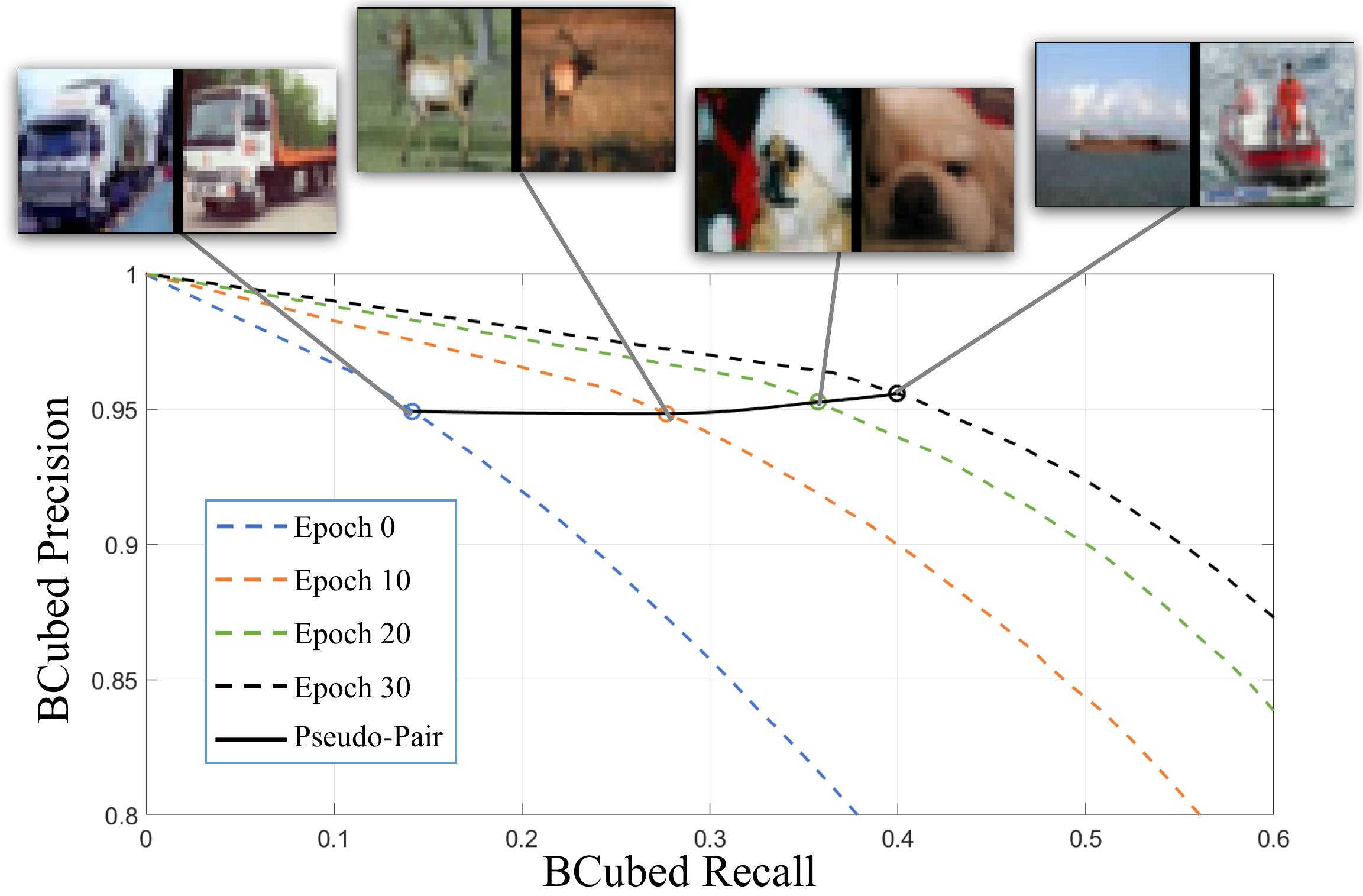}
	\caption{BCubed precision and recall curves~\cite{amigo2009comparison} for the pseudo-graphs of various epochs on CIFAR-10. These circle points on the lines correspond to the fixed pseudo-graph threshold $0.95$ in experiments.
	} 
	\label{fig:ROC}
	\vspace{-4mm}
\end{figure}

\noindent
\textbf{BCubed Precision and Recall of Pseudo-graph.}
BCubed~\cite{amigo2009comparison} is a metric to evaluate the quality of partitions in clustering.
We validate that our method can learn better representation in a progressive manner by using the BCubed~\cite{amigo2009comparison} precision and recall curves, which are computed based on the pseudo-graphs of different epochs in Figure~\ref{fig:ROC}.
%------------------------------------------------
%Our method can learn better representation in a progressive manner. 
%To validate this, we present the ROC curves for the pseudo-graphs of different epochs in Figure~\ref{fig:ROC}.
It is obvious that with the increasing of epochs, the precision of the pseudo-graph becomes much better, which will improve the clustering performance in return.

\begin{figure}[t]
	\centering 
	\subfigure[\scriptsize{Distribution of the largest probability}]{ 
		\label{fig:distribution} %% label for second subfigure 
		\includegraphics[width=0.5\linewidth]{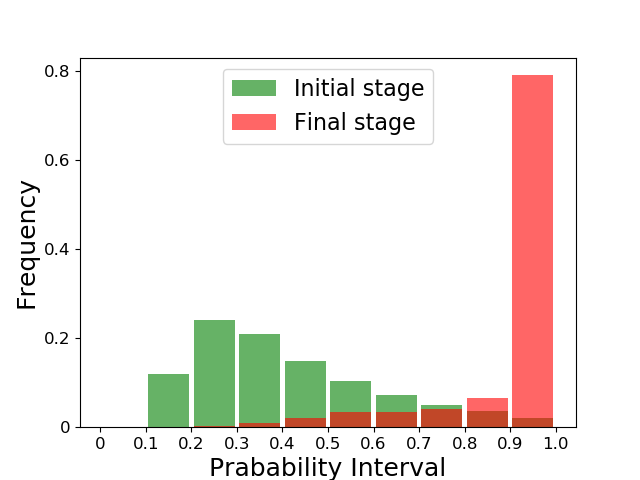}}
	\hspace{-4mm} 
	\subfigure[\scriptsize{Influence of $thres_2$}]{ 
		\label{fig:thres2_b} %% label for second subfigure 
		\includegraphics[width=0.45\linewidth]{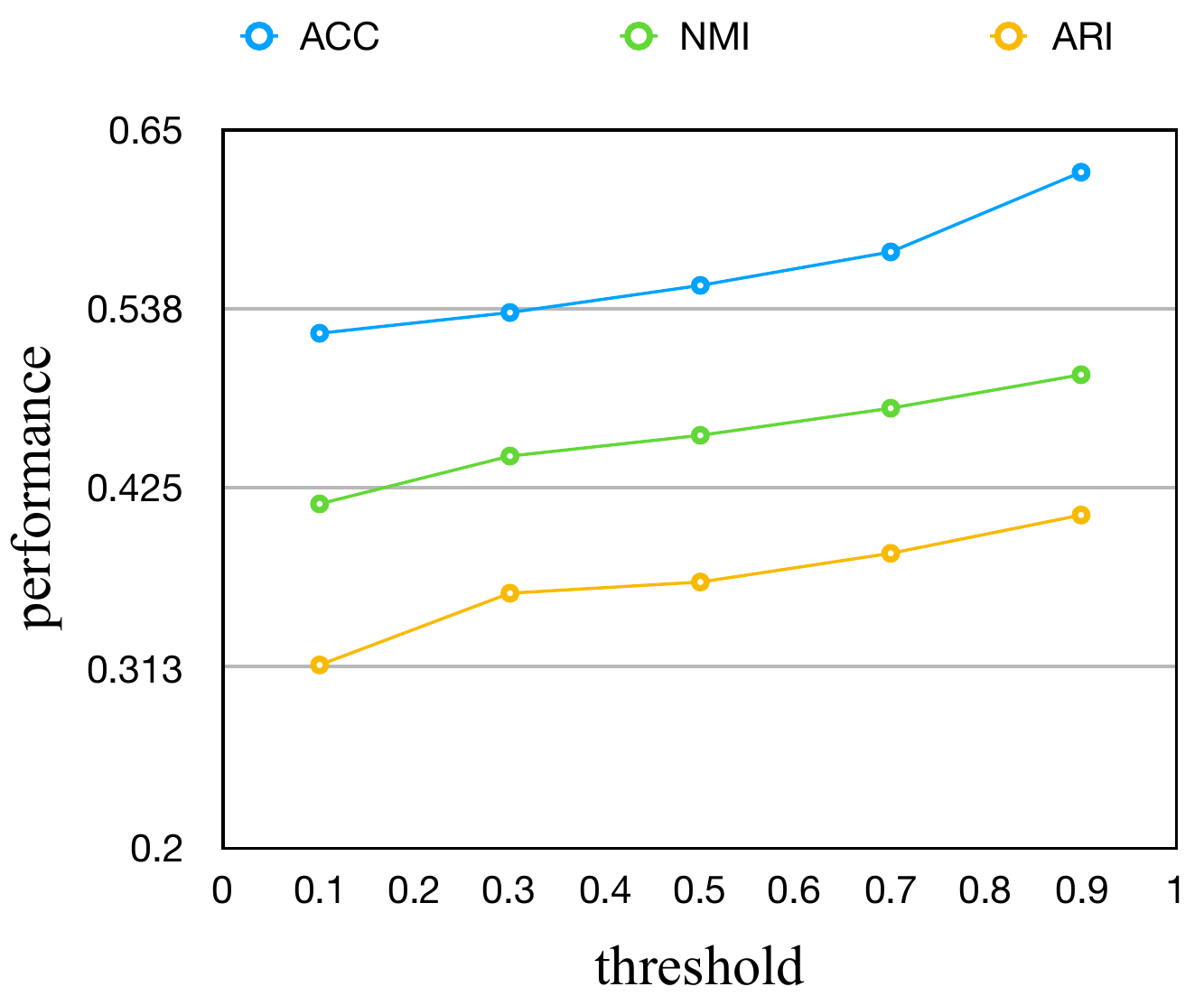}} 
	\caption{The distribution of the largest probability in all prediction features and the influence of threshold for highly-confident pseudo-label on the CIFAR-10 dataset.
	} 
	\label{fig:thres2}
	\vspace{-3.5mm}
\end{figure}

\noindent
\textbf{Statistics of Prediction Features.}
%------------------------------------------------
According to Claim~\ref{claim_one_hot}, the ideal prediction features have the one-hot property, so that we can use the highly-confident pseudo-label to guide the training.
To verify it, we compare the distribution of the largest prediction probability between the initial stage and the final stage.
The results on the CIFAR-10 dataset is presented in Figure~\ref{fig:distribution}.
For the CIFAR-10 dataset, the largest probability $p$ is in the range of $[0.1, 1]$. We count the probability in nine disjoint intervals, such as $[0.1, 0.2]$, $[0.2, 0.3]$, $\cdots$, and $[0.9, 1]$.
We can see that in the initial stage, less than $10\%$ of all samples have the probability that is larger than $0.7$, while after training, nearly $80\%$ of all samples have the probability that is larger than $0.9$.
The above results imply that the largest probability tends to be $1$, and others tend to be $0$, which is consistent with our Claim~\ref{claim_one_hot}.
\par

\noindent
\textbf{Influence of Thresholds.}
%------------------------------------------------
%------------------------------------------------------------------
%------------------------------------------------------------------
In Figure~\ref{fig:thres2}, we test the influence of threshold to select highly-confident pseudo-label for training.
We can see that with the increase of threshold, the performance also increases.
The reason is that with low threshold, some incorrect pseudo-label will be adopted for network training, which will affect the performance.
So it is important to set relatively high threshold to select highly-confident pseudo-label for supervision.
\section{Conclusions}
%------------------------------------------------
For deep unsupervised learning and clustering, we propose the DCCM to learn discriminative feature representation by mining comprehensive correlations.
Besides the correlation among different samples, we also make full use of the mutual information between corresponding features, local robustness to small perturbations, and their intercorrelations.
We conduct extensive experiments on several challenging datasets and two different tasks to thoroughly evaluate the performance. DCCM achieves significant improvement over the state-of-the-art methods.
\section*{Acknowledgment}
%\noindent \textbf{Acknowledgment}\\
The work of Z.~Lin was supported by 973 Program of China (grant no. 2015CB352502), NSF of China (grant nos. 61625301 and 61731018), Qualcomm, and Microsoft Research Asia. 
The work of H.~Zha was supported by the National Key Research and Development Program of China (grant no. 2017YFB1002601) and National Natural Science Foundation of China (grant nos. 61632003 and 61771026).
{\small
	\bibliographystyle{ieee_fullname}  
	\bibliography{egbib.bib}
}
\clearpage
%------------------------------------------------------------------ 
\section{Supplementary Material}
%------------------------------------------------------------------ 
%------------------------------------------------------------------ 
\subsection{Proof of Lemma~$1$ and Claim~$1$}
%------------------------------------------------------------------ 

\noindent\emph{Proof of Lemma~$1$}:
Since $\omega(e_i)\neq \omega(e_j)$ for $\forall i\neq j$, there exists a strongly increasing sequence of weights $\{\omega_1, \omega_2, \cdots, \omega_{\frac{N(N+1)}{2}}\}$, and we can remove edges from $G$ in the order from smallest weight to largest by increasing threshold $t$. This action would either increase the current partition number $n$ to $n+1$ or remain it unchanged. At the beginning of the process we have $1$ partition and at the end of the process we have $N$ partitions. Since $1\le K\le N$, there exists a $K$ partition in the process.

\hfill \qedsymbol
\vspace{5mm}

\noindent\emph{Proof of Claim~$1$}:
Select samples $\mathbf{x}^1,\mathbf{x}^2, \cdots,\mathbf{x}^K$ from partition $P^1,P^2,\cdots,P^K$, denote the cosine similarity matrix of their corresponding optimal features $f_{\theta*}(\mathbf{x}_1),f_{\theta*}(\mathbf{x}_2),\cdots,f_{\theta*}(\mathbf{x}_K)$ as $\mathbf{S}$, and $\mathbf{S}$ equals to its $K$ partitions pseudo graph $\mathbf{W}$, which is an identity matrix. Denote $f_{\theta*}(\mathbf{x}_i)$ as $[z_i^1,z_i^2,\cdots,z_i^K]$, where $z_{i}^{k} $ denotes the $k$-th element of the vector $\mathbf{z}_{i}$.

The set $\{z_1^1, z_1^2, \cdots, z_1^K, \cdots, z_K^1, \cdots,z_K^K\}$ can only have no more than $K$ positive elements, otherwise, according to Pigeonhole principle, there exists a $k$ that $z_{i}^k=z_{j}^k$ and $\cos(\mathbf{z}_{i},\mathbf{z}_{j})>0$, which is contradicted to $\mathbf{S}_{ij}=0$.

On the other hand, for the output of a softmax layer, every vector has at least one positive entry. Therefore, every vector has and only has one positive element that equals to $1$.

\hfill \qedsymbol

%------------------------------------------------------------------ 
\begin{table*}[!t]\label{tab:arch}
	\caption{Network architecture for various datasets we used in experiments.}
	\centering
	\renewcommand\arraystretch{1.1}
	\setlength{\tabcolsep}{7pt}
	\begin{tabular}{l|c|c|c|}
		\cline{2-4}
		& \begin{tabular}[c]{@{}c@{}}CIFAR-10 / CIFAR-100\\ 32$\times$32$\times$3\end{tabular}                                                                                                                                                                 & \begin{tabular}[c]{@{}c@{}}Tiny-ImageNet\\ 64$\times$64$\times$3\end{tabular}                                                                                                                                                                                                                                                                  & \begin{tabular}[c]{@{}c@{}}ImageNet-10/ImageNet-dog-15/STL-10\\ 96$\times$96$\times$3\end{tabular}                                                                                                                                                                                                                                                                              \\ \cline{2-4} 
		& \begin{tabular}[c]{@{}c@{}}3$\times$3 conv. 64 BN ReLU\\ (S)$\quad$3$\times$3 conv. 64 BN ReLU\\ 2$\times$2 MaxPooling with stride 2\\ 3$\times$3 conv. 128 BN ReLU\\ 2$\times$2 MaxPooling with stride 2\\ 3$\times$3 conv. 256 BN ReLU\\ 4$\times$4 AvgPooling with stride 2\end{tabular} & \begin{tabular}[c]{@{}c@{}}5$\times$5 conv. 64 BN ReLU\\ 5$\times$5 conv. 64 BN ReLU\\ 4$\times$4 MaxPooling with stride 4\\  (S)$\quad$3$\times$3 conv. 128 BN ReLU\\ 3$\times$3 conv. 128 BN ReLU\\ 4$\times$4 MaxPooling with stride 4\\ 1$\times$1 conv. 256 BN ReLU\\ 2$\times$2 AvgPooling with stride 2\end{tabular} & \begin{tabular}[c]{@{}c@{}}5$\times$5 conv. 64 BN ReLU\\ 5$\times$5 conv. 64 BN ReLU\\ 4$\times$4 MaxPooling with stride 4\\  (S)$\quad$3$\times$3 conv. 128 with BN ReLU\\ 3$\times$3 conv. 128 BN ReLU\\ 4$\times$4 MaxPooling with stride 4 \\ 1$\times$1 conv. 256 with BN ReLU\\ 4$\times$4 AvgPooling with stride 4\end{tabular} \\ \cline{2-4} 
		& \begin{tabular}[c]{@{}c@{}}(D)$\quad$Linear(256, 64) BN ReLU\\ Linear(64, c)\\ SoftMax\end{tabular}                                                                                                                                        & \begin{tabular}[c]{@{}c@{}}(D)$\quad$Linear(256, 256) BN ReLU\\ Linear(256, c)\\ SoftMax\end{tabular}                                                                                                                                                                                                                                & \begin{tabular}[c]{@{}c@{}}(D)$\quad$Linear(256, 64) BN ReLU\\ Linear(64, c) \\ SoftMax\end{tabular}
		\\ \cline{2-4} 
	\end{tabular}
\end{table*}
%------------------------------------------------------------------ 

%------------------------------------------------------------------ 
\subsection{Defenitions of Metrics}
%------------------------------------------------------------------ 

We introduce the following three standarded metrics we used to evaluate our model:
\begin{itemize}
	\item Normalized Mutual Information~(NMI): Let $C$ and $C'$ denote the predicted partition and the ground truth partition respectively, the NMI metric is calculated as:
	\begin{equation}
		\small
		\operatorname{NMI}(C,C')=\frac{\sum^K_{i=1}\sum^S_{j=1}|C_i\cap C'_j|\log \frac{N|C_i\cap C'_j|}{|C_i||C'_j|}}{\sqrt{(\sum^K_{i=1}|C_i|\log\frac{C_i}{N})(\sum^S_{j=1}|C'_j|\log\frac{C'_j}{N})}}.
	\end{equation}

	\item Adjusted Rand Index~(ARI): Given a set $S$ of $n$ elements, and two groupings or partitions~(e.g. clustering results) of these elements with $r$ and $s$ groups, namely $X=\{X_{1},X_{2},\ldots ,X_{r}\}$ and $Y=\{Y_{1},Y_{2},\ldots ,Y_{s}\}$, the overlap between $X$ and $Y$ can be summarized in a contingency table $\left[c_{ij}\right]$, where each element $c_{ij}$ denotes the number of objects in common between $X_{i}$ and $Y_{j}$: 
	\begin{equation}
		c_{ij}=|X_{i}\cap Y_{j}|.
	\end{equation}
	The contingent table is of the following shape:
	$$
	{{\begin{array}{c|cccc|c}{{} \atop X}\!\diagdown \!^{Y}&Y_{1}&Y_{2}&\ldots &Y_{s}&{\text{Sums}}\\\hline X_{1}&c_{11}&c_{12}&\ldots &c_{1s}&a_{1}\\X_{2}&c_{21}&c_{22}&\ldots &c_{2s}&a_{2}\\\vdots &\vdots &\vdots &\ddots &\vdots &\vdots \\X_{r}&c_{r1}&c_{r2}&\ldots &c_{rs}&a_{r}\\\hline {\text{Sums}}&b_{1}&b_{2}&\ldots &b_{s}&\end{array}}}
	$$
	and ARI is defined by:
	\begin{equation}
		\operatorname{ARI} ={\frac  {\sum _{ij}{\binom {n_{ij}}{2}} -[\sum _{i}{\binom {a_{i}}{2}}\sum _{j}{\binom {b_{j}}{2}}]/{\binom {n}{2}} }{ {{\frac {1}{2}}[\sum _{i}{\binom {a_{i}}{2}}+\sum _{j}{\binom {b_{j}}{2}}]} - {[\sum _{i}{\binom {a_{i}}{2}}\sum _{j}{\binom {b_{j}}{2}}]/{\binom {n}{2}}} }}.
	\end{equation}
	
	\item Accuracy~(ACC): Suppose the clustering algorithm is tested on $N$ samples. For a sample $\mathbf{x}_i$, we denote its cluster label as $r_i$ and its ground truth as $t_i$. The clustering accuracy is defined by:
	\begin{equation}
		\operatorname{ACC}(R, T) = \frac{\sum^N_{i=1}\delta(t_i,\operatorname{map}(r_i))}{N},
	\end{equation}
	where
	\begin{equation}
		\delta(a,b) = 
		\left\{
		\begin{aligned}
			&1, \ \  \text{if} \ \ a=b,\\
			&0, \ \ \text{otherwise,} \\
		\end{aligned}
		\right.
	\end{equation}
	and function $\operatorname{map}(x)$ denotes the best permutation mapping function gained by Hungarian algorithm~\cite{cai2005document}.
\end{itemize}

%Example code for running DCCM can be found at REDACTED.
%------------------------------------------------------------------ 
%------------------------------------------------------------------------
\begin{table*}[!ht]
	\caption{Result comparison under the same architecture~(except the last layer of RotNet) on CIFAR-10/100. '$<$' denotes 'less than'.}
	%\scriptsize
	\centering
	\setlength{\tabcolsep}{4pt}
	\renewcommand\arraystretch{1.2}
	\label{tab:compare}
	\begin{tabular}{|c|c|c|c|c|c|c|c|c|}
		\hline
		\multirow{2}{*}{} & \multicolumn{4}{c|}{CIFAR-10}                                                                                                 & \multicolumn{4}{c|}{CIFAR-100}                                                                                                \\ \cline{2-9} 
		&  NMI  & \begin{tabular}[c]{@{}c@{}}Clustering\\ ACC\end{tabular} & ARI & \begin{tabular}[c]{@{}c@{}}Classify\\ ACC\end{tabular} &  NMI   & \begin{tabular}[c]{@{}c@{}}Clustering\\ ACC\end{tabular} & ARI & \begin{tabular}[c]{@{}c@{}}Classify\\ ACC\end{tabular} \\ \hline
		RotNet~\cite{gidaris2018unsupervised}              &   0.316         &    0.389     & 0.139    &  0.755                                                         &       0.208                                              & 0.225    & 0.070    &   0.453                                                     \\ \hline
		%DeepCluster       &      0.095          & 0.235  & 0.055  &                                                        &    0.056 & 0.121     &  0.021    &                                                        \\ \hline
		DeepCluster~\cite{caron2018deep}       &      $<$0.3          & $<$0.3  & $<$0.1  &     $<$0.75                                                   &    $<$0.2 & $<$0.2     &  $<$0.07    &    $<$0.45                                              \\ \hline
		DAC~\cite{chang2017deep}               &     0.439  & 0.514    &  0.335   &    0.787                                                    &       0.228           &  0.254  &  0.121   &   0.485                                                     \\ \hline
		DCCM~(ours)              &     \textbf{0.496}                                                     &    \textbf{0.623} & \textbf{0.408}    &                        \textbf{0.818}                               &      \textbf{0.285}                                                        &  \textbf{0.327}       &  \textbf{0.173}      &                       \textbf{0.512}                                     \\ \hline
	\end{tabular}
\end{table*}
%-------------------------------------------------------------------------
%------------------------------------------------------------------ 
\subsection{Compared Methods}
%------------------------------------------------------------------ 
For clustering, we adopt both traditional methods and deep learning based methods, including K-means, spectral clustering~(SC)~\cite{zelnik2005self}, agglomerative clustering~(AC)~\cite{gowda1978agglomerative}, the nonnegative matrix factorization~(NMF) based clustering~\cite{cai2009locality}, auto-encoder~(AE)~\cite{bengio2007greedy}, denoising auto-encoder~(DAE)~\cite{vincent2010stacked}, GAN~\cite{radford2015unsupervised}, deconvolutional networks~(DECNN)~\cite{zeiler2010deconvolutional}, variational auto-encoding~(VAE)~\cite{kingma2013auto}, deep embedding clustering~(DEC)~\cite{2016ICMLDEC}, jointly unsupervised learning~(JULE)~\cite{yang2016joint}, and deep adaptive image clustering~(DAC)~\cite{chang2017deep}.

For classification task, we compare DCCM against several unsupervised feature learning methods, including variational auto-encoder~(VAE)~\cite{kingma2013auto}, adversarial auto-encoder~(AAE)~\cite{makhzani2015adversarial}, BiGAN~\cite{donahue2016adversarial}, noise as targets~(NAT)~\cite{bojanowski2017unsupervised}, and deep infomax~(DIM)~\cite{hjelm2018learning}.

%%%%%%%%% BODY TEXT
%------------------------------------------------------------------ 
\subsection{Architechtures Details}
%------------------------------------------------------------------ 
In Table~\ref{tab:arch}, we present the architectures for different datasets.
\par
For CIFAR-10/CIFAR-100~\cite{krizhevsky2009cifar}, we set $4$ conv layers and $3$ pooling layers, followed with $2$ fully-connected layers. Batch Normalization~\cite{ioffe2015batch} and ReLU are used on all hidden layers.
%, with \emph{track\_running\_stats} set to be False. 
The output features after the second conv layer~(S for shallow) and the first fc layer~(D for deep) are used to compute the mutual information~($\operatorname{MI}$) loss, concatenated as the input of discriminator. For other datasets, such as Tiny-ImageNet~\cite{deng2009imagenet} and STL-10~\cite{coates2011analysis}, we set $5$ conv layers instead of $4$.
Due to their larger input size, we use the feature maps after the third conv layer as S. For all experiments, the output was a $class\_num$ dimensional vector.

%------------------------------------------------------------------------
\subsection{Comparison Under the Same Architecture}
%------------------------------------------------------------------------
In Table~\ref{tab:compare}, we present the additional comparisons using the same network. On CIFAR-10/100,
DeepCluster does not work well based on its released official PyTorch code. 
DAC has similar performance with that in their paper.
Our DCCM achieves the best results.

Please note that we only use a simple shallow version of AlexNet in the paper,
and our results are much better than the best reported results of other methods.

Besides, our algorithm is relatively efficient. On CIFAR-100, it costs 19 hours for training on a single GTX 1080Ti GPU.
Multiple GPU cards and better GPU can improve this.

%------------------------------------------------------------
\subsection{Sampling Strategy}
%------------------------------------------------------------
%------------------------------------------------------------
\begin{table}[!ht]
	\renewcommand\arraystretch{1.3}
	\caption{Classification accuracy of different pair-sampling strategies on CIFAR-10.}
	\setlength{\tabcolsep}{1pt}
	\begin{tabular}{|c|c|c|}		
		\hline  & Methods & Classification ACC(Y64) \\ \hline
		V1 & 	nearest pos + random* neg 	& 0.744	\\ \hline
		V2 & 	nearest pos + farthest neg 	& 0.713 \\ \hline
		V3 & 	random* pos + random* neg 	& 0.731 \\ \hline
		V4 & 	top-n pos + random* neg 	& 0.698 \\ \hline
	\end{tabular}
	\label{tab:sampling}
\end{table}
%------------------------------------------------------------
The experiment result corresponding to the analysis in line $836$-$843$ is listed in Table~\ref{tab:sampling}.
We tried four strategies to fetch positive and negative pairs from pseudo-graph $\mathbf{W}$, and the terms used in the table refer to:
\begin{itemize}
	\item nearest means that for each sample, we select its nearest sample from the minibatch to construct a positive pair, while farthest means taking the farthest one to construct a negative pair.
	\item random* means that we randomly take a positive sample that satisfies $W_{ij}=1$ as a positive pair or a negative sample that satisfies $W_{ij}=0$ as a negative pair.
	\item top-$n$ pos means that we select the top $n$ confident pairs from the graph $W$ to construct positive pairs.
\end{itemize}  For each strategy, we take $n$ positive pairs and $n$ negative pairs into account, where $n$ is our batch size. This is to make sure that the computational complexity of each approach is nearly the same for fair comparison, while we also have explored more costly approaches and find that the improvement is negligible.  

To clearly illustrate how $L_{MI}$ is effected, here we set a fixed model trained with only $\widehat{L_{\operatorname{PG}}}+\widehat{L_{\operatorname{PL}}}$. Then with the pseudo-graph $\mathbf{W}$ generated by it, we train a new model using only $L_{\operatorname{MI}}$ from scratch. It can be concluded that the positive pairs are sensitive to noise since strategy V1 achieves better results than V3, and harder negative pairs are beneficial for training as strategy V1 also achieves better results than V2.
Besides, we also notice the importance of uniform sampling within the minibatch, as the top-n pairs in V4 has higher confidence than that in V1, but the training collapses since only part of samples in the batch are included in the top-n strategy.

\end{document}